\documentclass[letterpaper,10pt,conference]{IEEEtran}
\IEEEoverridecommandlockouts

\usepackage[acronym]{glossaries}
\newacronym{tr}{TR}{teach-and-repeat}
\newacronym{vtr}{VTR}{visual teach-and-repeat}
\newacronym{ugv}{UGV}{Unmanned Ground Vehicle}

\usepackage{fontawesome}
\usepackage{pifont}
\usepackage{stix}
\usepackage{combelow}

\usepackage{hyperref}
\newcommand\rurl[1]{%
\texttt{\href{http://#1}{\nolinkurl{#1}}}
}

\usepackage{amsfonts}

\usepackage{cite}

\usepackage[labelformat=simple]{subcaption}

\usepackage{algorithm}
\usepackage{algpseudocode}

\usepackage{cleveref}
\crefname{table}{Tab.}{Tabs.}
\crefname{figure}{Fig.}{Figs.}
\crefname{section}{Sec.}{Secs.}
\crefname{equation}{Eq.}{Eqs.}

\usepackage{graphicx}

\usepackage[binary-units]{siunitx}
\usepackage{threeparttable}
\usepackage{booktabs}

\usepackage{tikz}
\usetikzlibrary{shapes,arrows}

\usepackage{multicol}

\usepackage{todonotes}

\usepackage{fancyhdr}
\fancypagestyle{withfooter}{

\fancyfoot[C]{\footnotesize Accepted to the IEEE ICRA Workshop on Field Robotics 2024}
}

\begin{document}

\title{
\huge \bf \textit{Watching Grass Grow}: Long-term Visual Navigation and Mission Planning for Autonomous Biodiversity Monitoring
}
\author{Matthew Gadd$^1$, Daniele De Martini$^1$, Luke Pitt$^3$, Wayne Tubby$^3$, Matthew Towlson$^3$, Chris Prahacs$^3$,\\ Oliver Bartlett$^3$, John Jackson$^4$, Man Qi$^4$, Paul Newman$^1$, Andrew Hector$^4$, Roberto Salguero-G\'omez$^4$, Nick Hawes$^2$\\
$^1$Mobile Robotics Group (MRG), $^2$Goal-Oriented Long-Lived Systems (GOALS),\\ $^3$Oxford Robotics Institute, 
$^4$Department of Biology, University of Oxford\\
\faEnvelope~\texttt{mattgadd@robots.ox.ac.uk}
\thanks{
This project was supported by EPSRC Programme Grant ``From Sensing to Collaboration'' (EP/V000748/1), the EPSRC IAA Technology Fund (0012338), a John Fell Fund, NERC grants (NE/M018458/1 and NE/X013766/1) and the SustainTech-3 testbed within a AWS gift-supported Human-Machine Collaboration Programme.
We would also like to thank the Conservator of Wytham Woods, Nigel Fisher and his staff, at the University of Oxford.
}
}
\maketitle

\begin{abstract}
We describe a challenging robotics deployment in a complex ecosystem to monitor a rich plant community.
The study site is dominated by dynamic grassland vegetation and is thus visually ambiguous and liable to drastic appearance change over the course of a day and especially through the growing season.
This dynamism and complexity in appearance seriously impact the stability of the robotics platform, as localisation is a foundational part of that control loop, and so routes must be carefully taught and retaught until autonomy is robust and repeatable.
Our system is demonstrated over a $6$-week period monitoring the response of grass species to experimental climate-change manipulations.
We also discuss the applicability of our pipeline to monitor biodiversity in other complex natural settings.
\end{abstract}

\thispagestyle{withfooter}
\pagestyle{withfooter}

\section{Introduction}
\label{sec:introduction}

\begin{figure}[!h]
\centering
\begin{subfigure}{0.55\columnwidth}
\includegraphics[width=\linewidth]{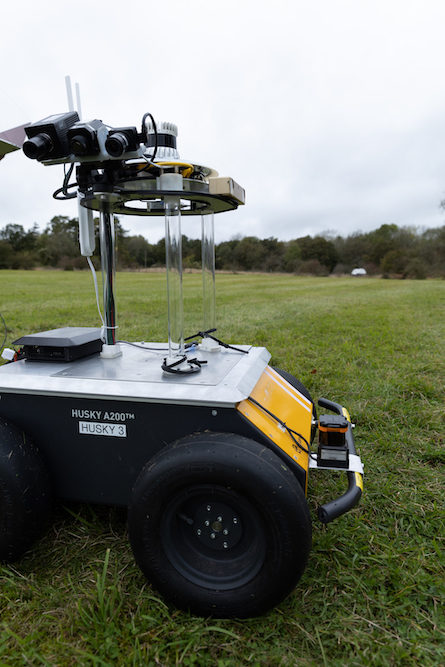}
\end{subfigure}
\begin{subfigure}{0.55\columnwidth}
\includegraphics[width=\linewidth]{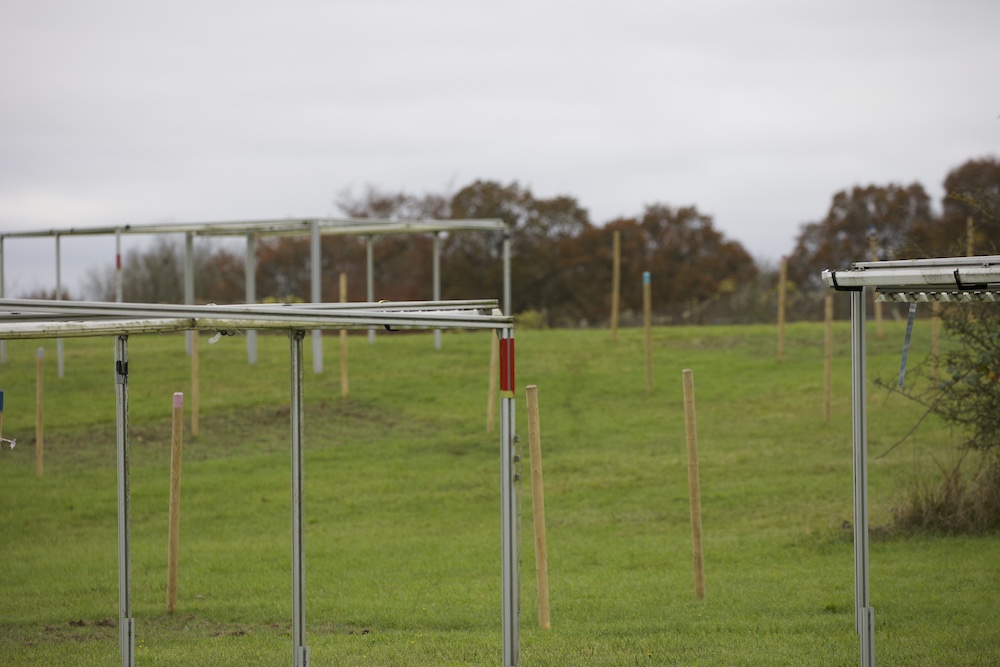}
\end{subfigure}
\caption{
\textit{ClearPath Husky} base platform (top) with \textit{Bumblebee2} stereo vision (facing forwards) camera used for mapping and localisation with \textit{Hokuyo} lasers (front bumper) for obstacle avoidance and safety curtain.
Pointing sideways are payload cameras, not used for navigation, including monocular, thermo and multi-spectral.
Plant-growing experiment enclosures (bottom), where rainwater is fed into four quadrants at different rates to investigate grass species' response to climate change.
}
\label{fig:husky-in-field}
\vspace{-20pt}
\end{figure}

Humans are having an immense impact on the natural world, to such an extent that scientists now recognise the current geological epoch as the Anthropocene~\cite{waters2016anthropocene}.
Climate change, habitat destruction, direct exploitation (e.g. poaching) and the facilitation of invasive species threaten over one million species with extinction before \num{2100}~\cite{pimm2014biodiversity}.
As a result, we need ambitious, effective, and flexible solutions to aid in monitoring and protecting biodiversity~\cite{kuhl2020effective}.
While remaining the essential core component of biodiversity research, direct monitoring of ecosystems by humans is expensive, inefficient, error-prone, and time-consuming at the spatial scales needed.
Therefore, autonomous data collection and processing (e.g. via robots and sensor networks) has the potential to rapidly improve the cost-effectiveness of biodiversity monitoring (e.g.~\cite{pollayil2023robotic}).

\begin{figure}[t]
\centering
\includegraphics[width=0.7\columnwidth]{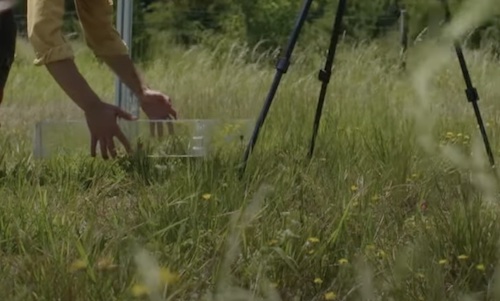}
\caption{
\textbf{Manual data collection} at one of the \num{40} grassland plots.
A researcher places a transparent box over the plants of interest, with interesting features to be drawn by hand.
}
\label{fig:manual_collection}
\vspace{-10pt}
\end{figure}

\begin{figure}
\centering
\includegraphics[width=0.8\columnwidth]{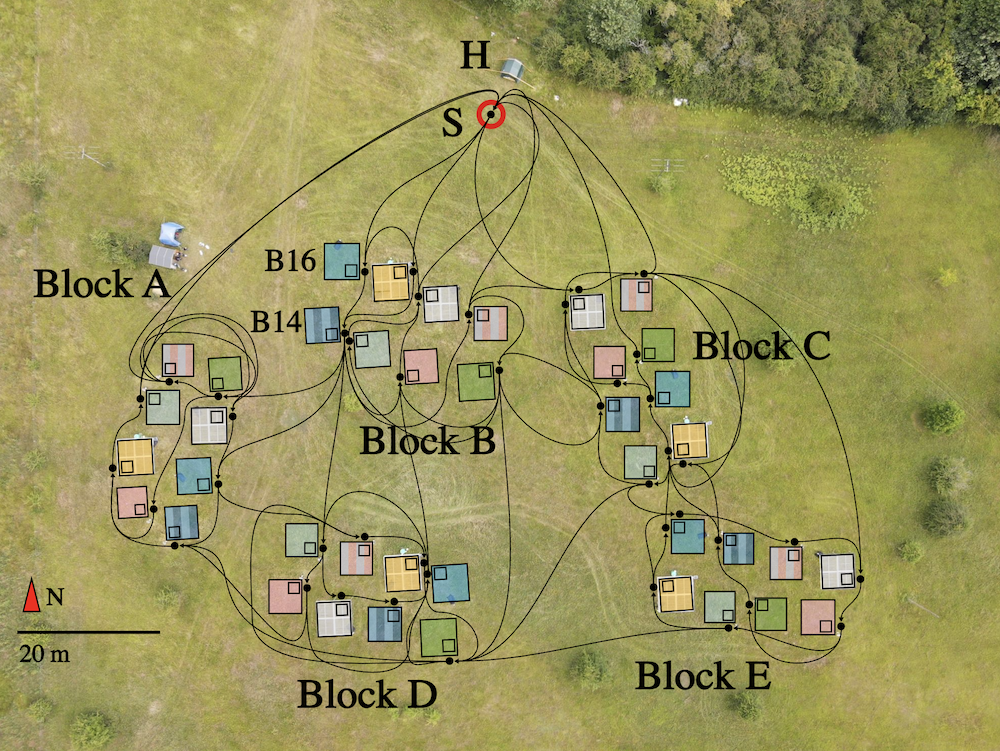}
\caption{
An aerial shot of the \textbf{\textit{Wytham Woods} experimental site}, with overlays indicating growing areas of interest for autonomous monitoring of complex, dynamic grasslands.
\textit{Wytham Woods} are a \num{423.8}-hectare biological \textit{Site of Special Scientific Interest} north-west of Oxford in Oxfordshire.
The topological map or ``supergraph'' (\cref{sec:planning}) for autonomous navigation is shown in black.
For example, in~\cref{fig:hx_map}, consider edges in the graph \texttt{S-B16} and \texttt{B16-B14} (all other node names are not shown, to avoid clutter).
Each edge also represents sequences of images or visual ``experiences'' (\cref{sec:dub4}).
Autonomous docking, charging, and data offload (\cref{sec:docking}) occurs at a tent (\cref{sec:hardware}) near node \texttt{H}.
\label{fig:hx_map}}
\vspace{-15pt}
\end{figure}

In this work, we present a system for biodiversity monitoring based on a combination of ground robots and multi-spectral imaging.
This system is as pictured in~\cref{fig:husky-in-field} currently installed at \textit{Wytham Woods}, a mixed ancient woodland near Oxford (UK) as per~\cref{fig:hx_map}, monitoring plant biodiversity changes at \num{40} grassland experimental permanent plots that belong to two global ecological monitoring networks: \textit{DroughtNet}\footnote{A global network to assess terrestrial ecosystem sensitivity to extreme drought, \rurl{droughtnet.weebly.com}} and \textit{DRAGNet}\footnote{A network to monitor disturbance and recovery across global grasslands, \rurl{dragnetglobal.weebly.com}}.
These \num{40} plots have been variously exposed (in one of the quadrants marked $\squarellquad$ in~\cref{fig:hx_map}) to eight different treatments: Control, Procedural Control, Drought,
Nutrient Addition,
Mechanistic Disturbance,
Nutrient Addition and Mechanistic Disturbance, and Nutrient Cessation.
These experimental treatments, which are being globally replicated at $>$60 other nodes in each network, simulate the effects of climate change and human disturbances on grassland ecosystems.
Currently, data collection is as per~\cref{fig:manual_collection} entirely manual.
In this paper, we describe field trials for a mobile robotic solution to automate this data collection.

\section{Related Work}
\label{sec:related_work}

Multiple applications of autonomous mobile robots have been developed to patrol sites of interest to monitor man-made~\cite{gadd2015framework,michal2024autoinspect} or natural~\cite{gadd2024oord} environments.

Most closely related to the system we have fielded in this paper are \glspl{ugv} monitoring plant growth.
In this area, in \cite{kumar2016smart} an autonomous gardening robotic vehicle is equipped with feature extraction algorithms and neural networks to identify and classify plant species to measure key gardening parameters such as temperature, humidity, and soil moisture.
Data collected by the vehicle's sensors are sent to a cloud storage platform for analysis and future predictions, with monitoring and control accessible via a website and an Android application.
In \cite{faryadi2019autonomous} an \gls{ugv} monitors multiple areas of greenhouses.
The system optimises the robot's trajectory for efficient monitoring, alongside a data collection and image processing algorithm enabling real-time detection of crop changes and construction of plant row maps.

In some cases, ground vehicles are not appropriate for the monitoring application.
In \cite{aucone2023drone}, an unmanned aerial vehicle (UAV, aka drone) samples environmental DNA from the outer branches of tree canopies for scalable detection of animal species in unreachable above-ground substrates.
%
Similarly, appropriate autonomous systems are important for monitoring marine environments.
For example, in \cite{jones2019autonomous}, marine autonomous systems equipped with acoustic, visual, and oceanographic sensors are used for environmental assessment and monitoring during decommissioning of oil and gas industry structures.
In \cite{kurkin2017autonomous}, an autonomous mobile robotic system is used to monitor the environmental state of coastal zones for marine natural disaster prediction, with a modular design adaptable to diverse conditions and incorporating various sensors and systems.

Some applications have been developed to monitor extra-terrestrial systems or harsh environments meant to simulate off-planet operations, such as \cite{curtis2017uk}, where areas with deposits laid down in water in a low-energy environment -- indicative of ancient habitability -- are located by an autonomous mobile robot, delivering a drilling tool to dig into the surface to acquire samples, and examining them with onboard sensors.

In some cases, patrolling a geographic area is not important for analysing the environment.
For instance, \cite{chaudhury2015computer} employs an automated system using laser scanners mounted on a robot arm to quantitatively measure 3D plant growth, towards non-invasive and non-destructive methods for plant phenotype analysis for crop production research.
Likewise, in \cite{wagele2022towards}, a network of automated multi-sensor stations is used to monitor species diversity, with each station including autonomous samplers for various organisms, audio recorders, sensors for volatile organic compounds, and camera traps.

\begin{figure}[t]
\centering
\includegraphics[width=0.9\columnwidth]{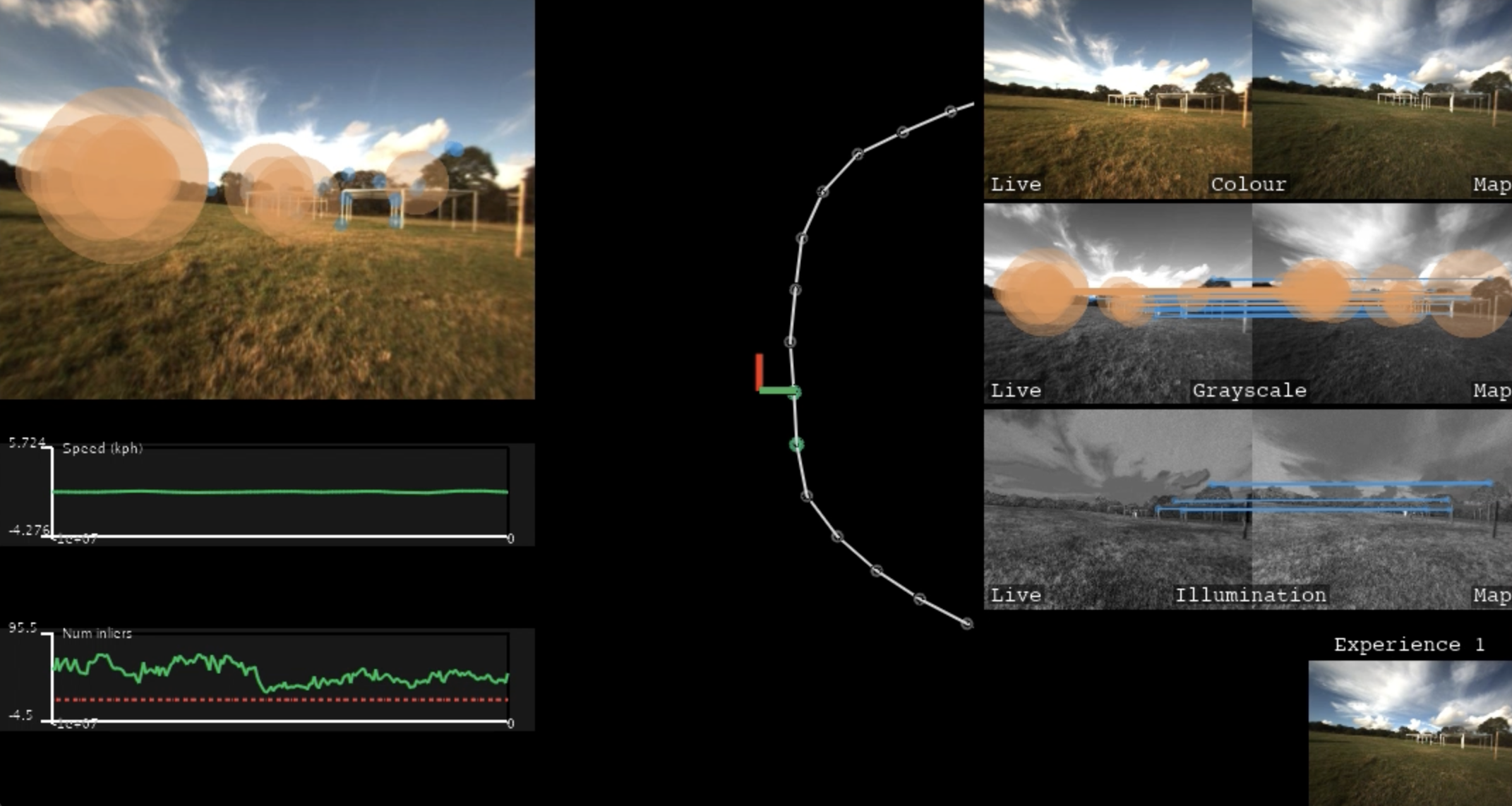}
\caption{
\textbf{Vision-based mapping and localisation}, based on~\cite{linegar2015work}.
The robot frame of reference is shown as the red and green axes, just to the left of a path previously built by an odometry chain (white).
Local ORB features~\cite{rublee2011orb} (orange and blue) are used to estimate precise pose with respect to the path.
The framework supports multi-experience localisation (bottom right), with just $1$ experience in this example.
Illumination-invariant image transformations~\cite{mcmanus2014shady} (right) are used for improve robustness to glare and shadows, etc.
}
\label{fig:dub4}
\vspace{-15pt}
\end{figure}

\section{Autonomous System Description}
\label{sec:system}

We use an \gls{ugv} with navigation sensors to enable robust autonomy around the site alongside payload sensors to automate the data-collection activities shown in~\cref{fig:manual_collection}.

\subsection{Vision-based localisation}
\label{sec:dub4}

Here, we use the vision-based mapping and localisation system described in~\cite{linegar2015work} to locate the robot within the world (\cref{fig:dub4}).
In this, the underlying representation of the world takes the form of a topometric ``experience graph'' (shown white) -- i.e. a graph, where nodes hold information about 3D landmarks, while edges contain relative poses with six degrees of freedom (6DoF), all obtained through visual odometry (VO)~\cite{nister2004visual}.
This VO pipeline performs joint optimisation of local landmarks and frame-to-frame relative poses derived from stereoscopic frames.
Coarse localisation is performed via a FAB-MAP search \cite{cummins2008fab} over bags of visual words~\cite{yang2007evaluating}, giving seed poses.
Fine localisation from these seed poses is then accomplished through breadth-first searches within local graph neighbourhoods.
This final step is successful when acquiring a stereo match with an adequate number of inliers.

\subsection{Teach-and-repeat}
\label{sec:vtr}

Our robot moves autonomously through \gls{vtr}~\cite{furgale2010visual}.
In \gls{tr}, a robot learns a path taught to it when piloted manually by a human operator; once taught, the robot repeats the path autonomously.
In our case, the taught paths (white) are recorded as a sequence of keyframes storing images, bags of visual words~\cite{yang2007evaluating}, image descriptors and point features.
Starting with a refined pose (\cref{sec:dub4}), the robot plans to follow the trajectory built of the chain of VO edges (white in~\cref{fig:dub4}) forwards.

These paths are taught over the full route network depicted (black) in~\cref{fig:hx_map}.
This is a complex path over many kilometres, and so a single data collection outing is not feasible.
\Cref{sec:mapping} therefore describes a mapping workflow/schedule to sufficiently capture our operational domain to allow good autonomy throughout.

\subsection{Topological mission-planning}
\label{sec:planning}

A na\"ive application of \gls{tr} would have the robot visiting each of the experimental permanent plots of~\cref{fig:hx_map} in exactly the order they were visited during data collection.
This approach of navigation, however, is undesirable, as the priority for visits is determined by the overall research study, the vagaries of weather, personal scheduling for access to the site, etc.

Therefore, alongside the experience graph, we also represent the environment with a coarser, higher-level graph designed specifically for mission scheduling.
This ``supergraph'' has nodes at each of the plant growing sites, each corresponding to multiple nodes or keyframes in the experience graph (\cref{sec:dub4,sec:vtr}), which are themselves at the termination or start of mapping runs (see below in~\cref{sec:mapping}).

Given a set of target nodes in the ``supergraph'' to capture data with the payload sensors (\cref{sec:sensors}), we currently solve a travelling salesman problem (TSP) and plan the shortest path through the graph to reach each node following the TSP tour. 
In the future, with historic traversal data over months of traversals, we also plan to enable probabilistic mission planning of the kind discussed in~\cite{lacerda2019probabilistic} to e.g. maximise the expected localisation performance during repeat phases.

\subsection{LiDAR safety curtain}
\label{sec:scs}

The robot's laser-based obstacle avoidance safety curtain is based on the \textit{Hokuyo} lasers shown in~\cref{fig:husky-in-field}.
These lasers are mounted with their beam sweeping horizontally (parallel to the ground).
The safety curtain sets speed limits and estop assertions by detecting obstacles in the LiDAR data.
To carry out this task, the robot defines objects as point groups exceeding a certain size with no internal gaps larger than another predefined size.

This safety curtain is crucial to the long-term, semi-unsupervised deployment of our platform at the \textit{Wytham Woods} site, as  \textit{Wytham Woods} -- while primarily a \textit{University of Oxford} Nature Conservation and experimental site -- are accessible by walking permit.
Additionally, wild animals are common on the site (badgers, rabbits, muntjacs, etc).
While we did not encounter any such participants during our experiment, collision avoidance is nevertheless crucial.

\subsection{Autonomous docking \& charging}
\label{sec:docking}

\begin{figure}
\centering
\includegraphics[width=0.6\columnwidth]{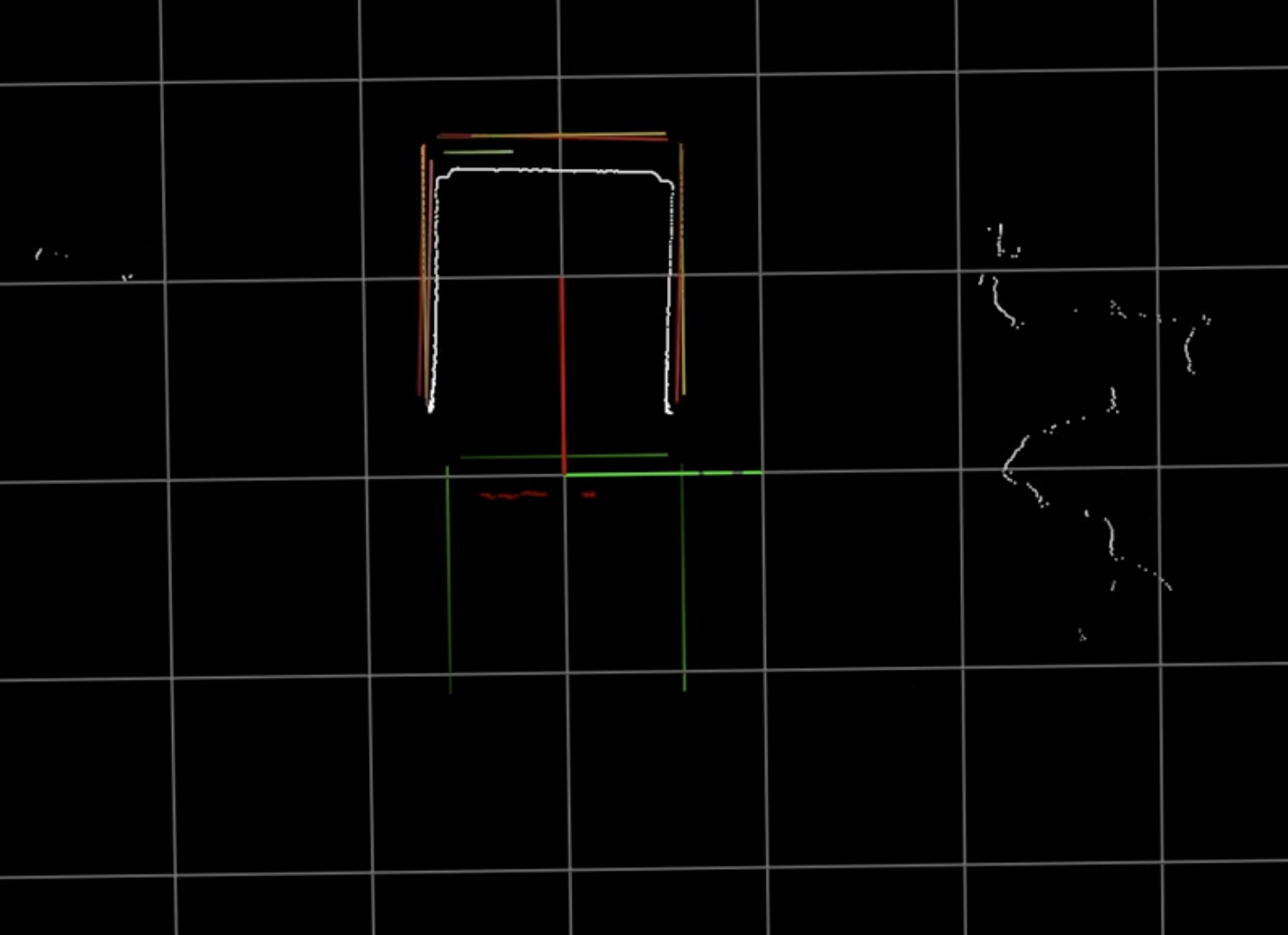}
\caption{
\textbf{Autonomous docking LiDAR perception system}.
The white dots are LiDAR returns. The red lines are detected line segments. The orange ones are the location of the predefined model that was chosen to best match the detected line segments. The green lines are the desired locations of the docking station layout after docking is complete -- i.e. the control reference signal.
}
\label{fig:docking}
\vspace{-15pt}
\end{figure}

Automatic docking and charging are depicted in~\cref{fig:docking}.
To allow for totally autonomous charging, we use the wireless \textit{Wibotic} charger (\rurl{wibotic.com/products/hardware}) and the docking functionality is based on the 2D LiDAR (\cref{fig:husky-in-field}).

We use a custom line fitting algorithm similar to RANSAC~\cite{derpanis2010overview} but deterministic, where domain-specific prior information (i.e. we know the shape of the dock) and previous model detections are exploited to optimise this search.
These line segments are matched against a predefined model.
The best match is chosen -- provided there is a valid match -- to determine the relative transformation of the dock to the robot.

Our docking controller is implemented as a simple state machine.
The controller has a concept of the ``runway'', which is a line segment which ends at the location in the dock the robot should end up at, and starts a short distance outside the dock.
If the robot is some distance away from the runway, it will drive towards it.
If the robot is close enough to the runway, then it will align its body with the runway by rotating on the spot.
Then, the distance to the end of the runway is fed into a PID controller to control speed and the angle to the runway is used to control rotation with a separate PID controller.
If the dock has not been detected in the past preset number of seconds the controller will have the robot rotate on the spot in the direction the dock was last seen.

\begin{figure}[t]
\centering
\begin{subfigure}{0.44\columnwidth}
\includegraphics[width=\linewidth]{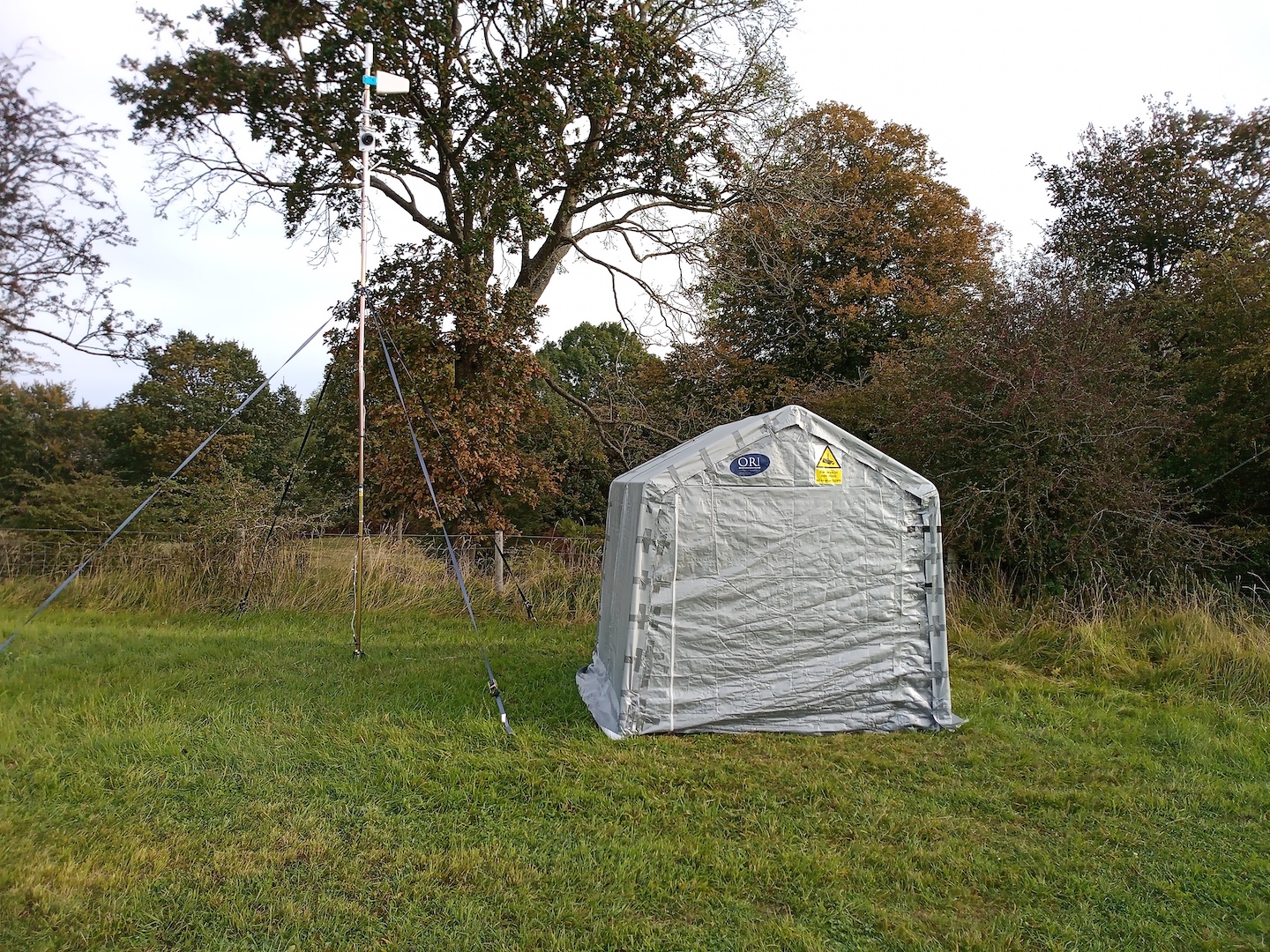}
\caption{\label{fig:tent}}
\end{subfigure}
\begin{subfigure}{0.44\columnwidth}
\includegraphics[width=\linewidth]{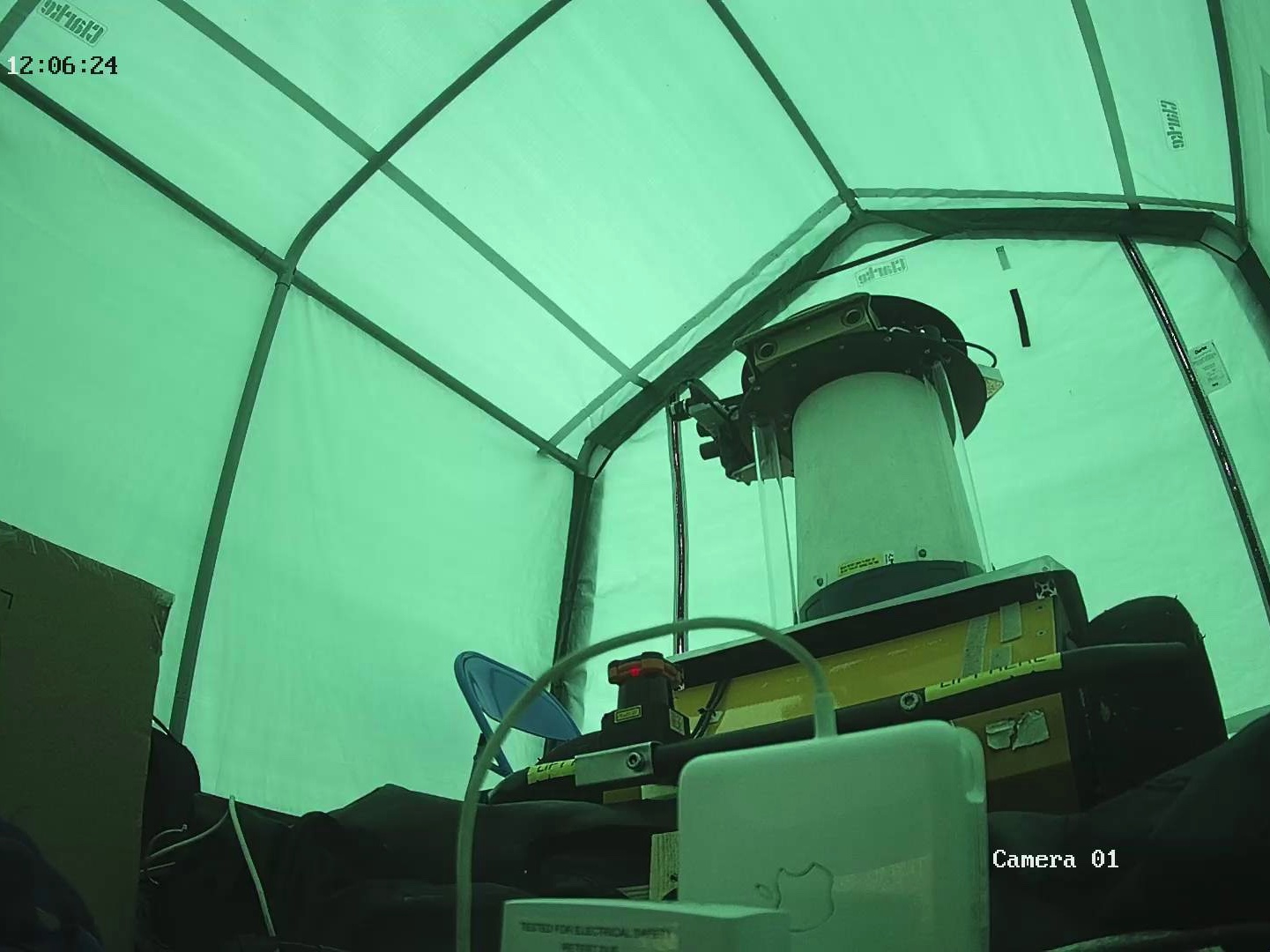}
\caption{\label{fig:husky_in_tent}}
\end{subfigure}
\caption{
\textbf{Charging and data offload centre}.
This corresponds to place/node \texttt{H} in~\cref{fig:hx_map}, with staging zone \texttt{S} just in front of the tent used to trigger controller handover from teach-and-repeat (\cref{sec:vtr}) to LiDAR-based docking (\cref{sec:docking}).
}
\label{fig:doghouse}
\vspace{-5pt}
\end{figure}

\subsection{Hardware considerations}
\label{sec:hardware}

In~\cref{fig:tent}, we show a weatherproof ``doghouse'' enclosure installed at the site, to allow for the robot to dock and charge overnight (see~\cref{sec:docking}), as well as offload data collected during the day (to make space on its internal hard drive).
To the left of the tent is a 4G mast which we use to stream images from a web camera inside the tent, an example of which is shown in
\cref{fig:husky_in_tent} with a view of the robot, and eventually data collected during the trials for processing and analysis.

Note that docking and charging are also essential for daylong autonomy, as the batteries used to power both the platform's motors and the robot's internal computers were limited to less than \SI{2}{\hour} under continuous operation (see~\cref{fig:autonomous_time} with the longest autonomous trial being \SI{1}{\hour} and \SI{45}{\minute}).

\subsection{Sensor suite}
\label{sec:sensors}

\begin{figure}[t]
\centering
\includegraphics[width=0.5\columnwidth]{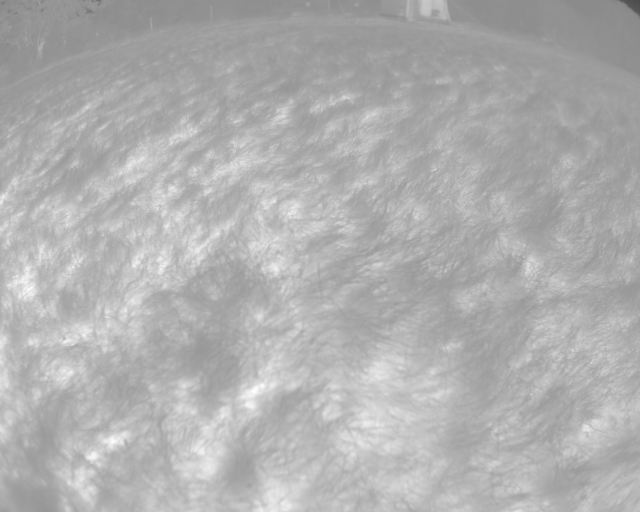}
\caption{
\textbf{Payload thermal camera} example image, useful to analyse plant growth.
}
\label{fig:thermal_frames}
\vspace{-15pt}
\end{figure}

\begin{figure*}[!h]
\centering
\begin{subfigure}{0.17\textwidth}
\includegraphics[width=\linewidth]{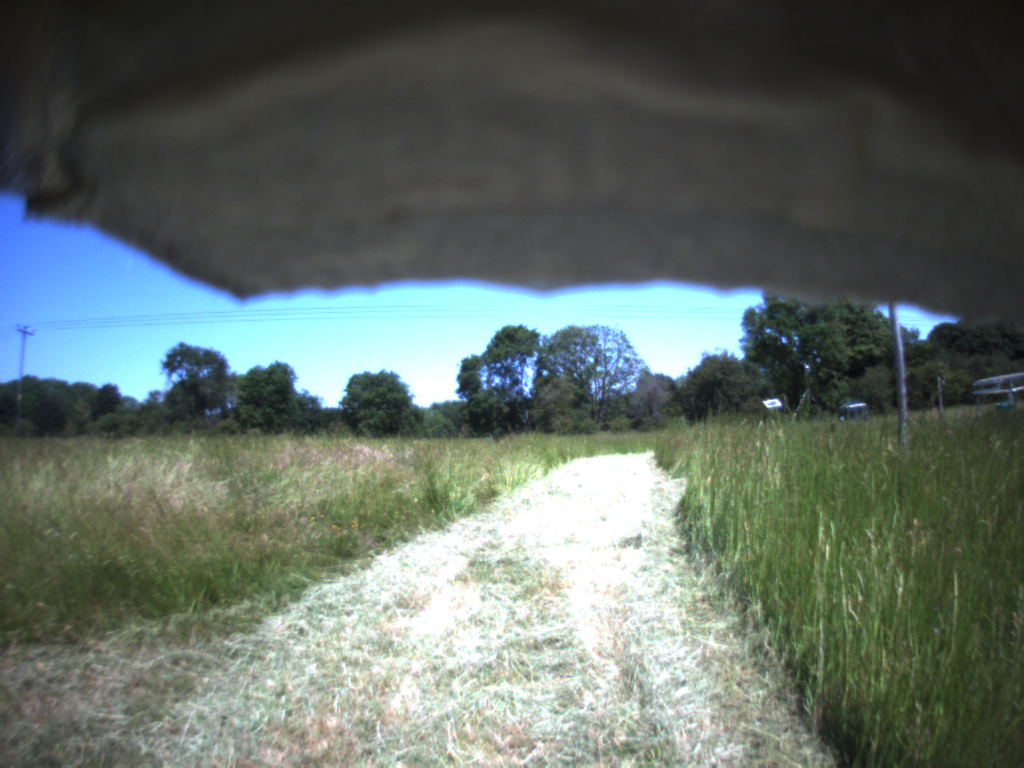}
\end{subfigure}
\begin{subfigure}{0.17\textwidth}
\includegraphics[width=\linewidth]{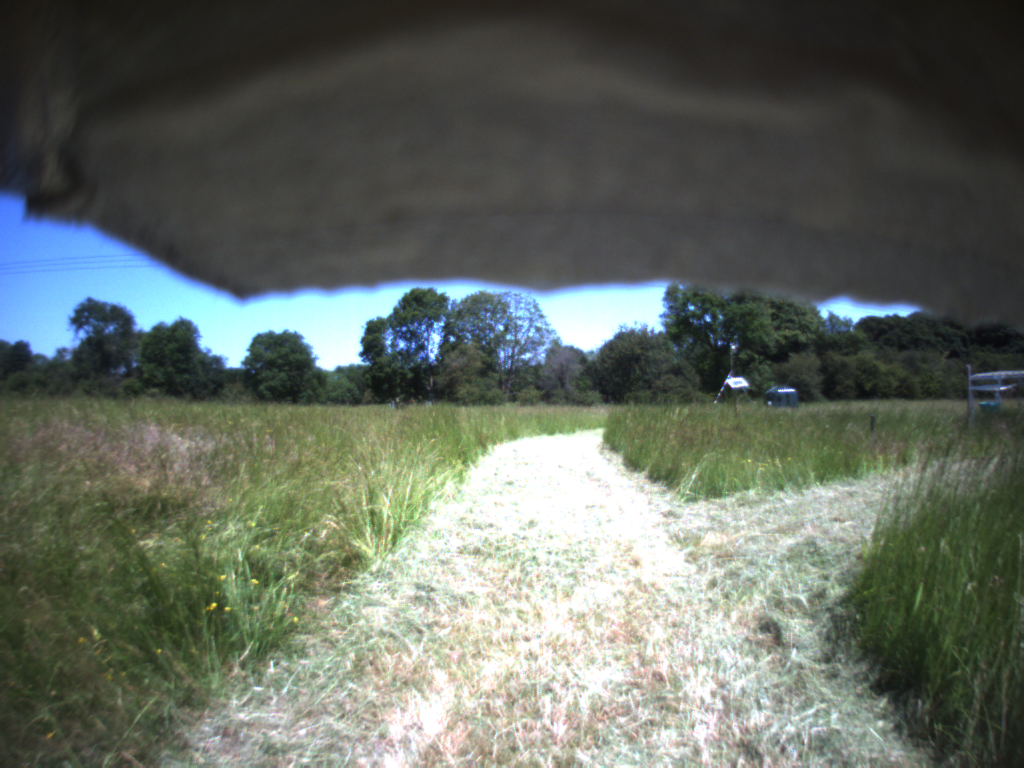}
\end{subfigure}
\begin{subfigure}{0.17\textwidth}
\includegraphics[width=\linewidth]{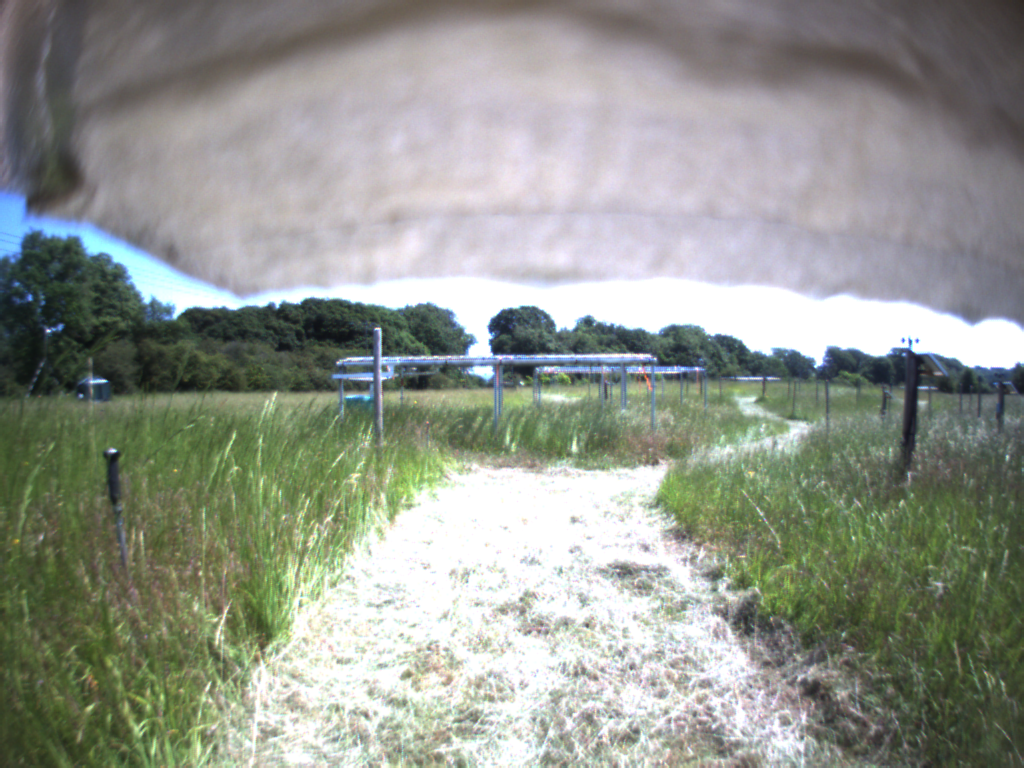}
\end{subfigure}
\begin{subfigure}{0.17\textwidth}
\includegraphics[width=\linewidth]{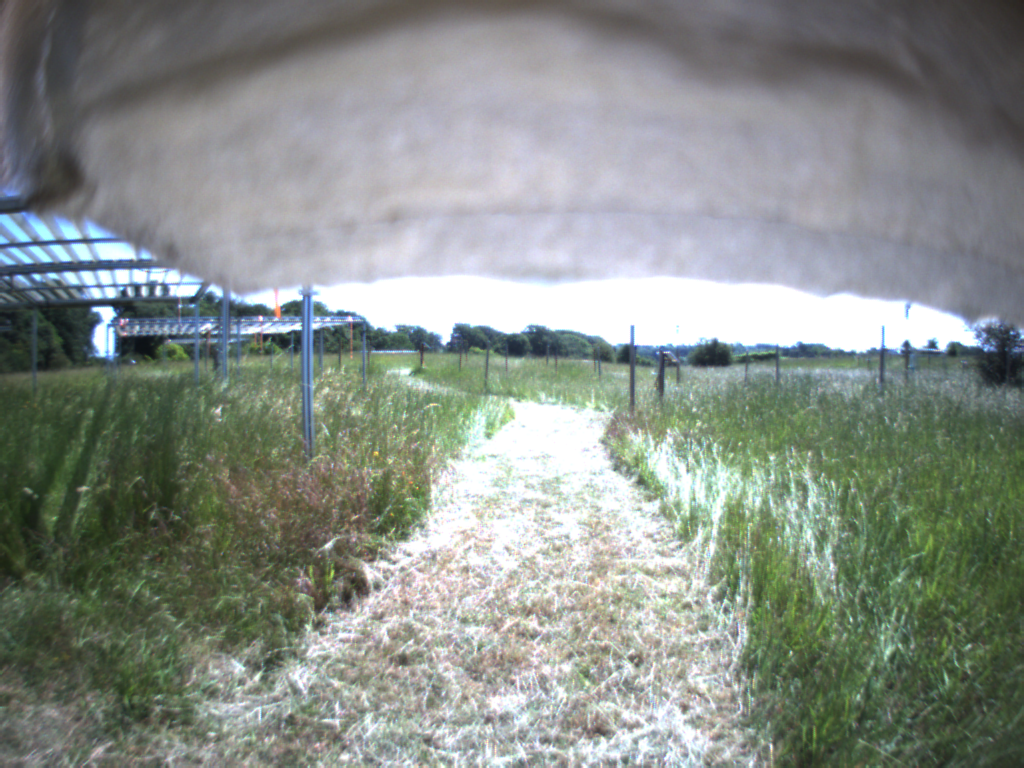}
\end{subfigure}
\begin{subfigure}{0.17\textwidth}
\includegraphics[width=\linewidth]{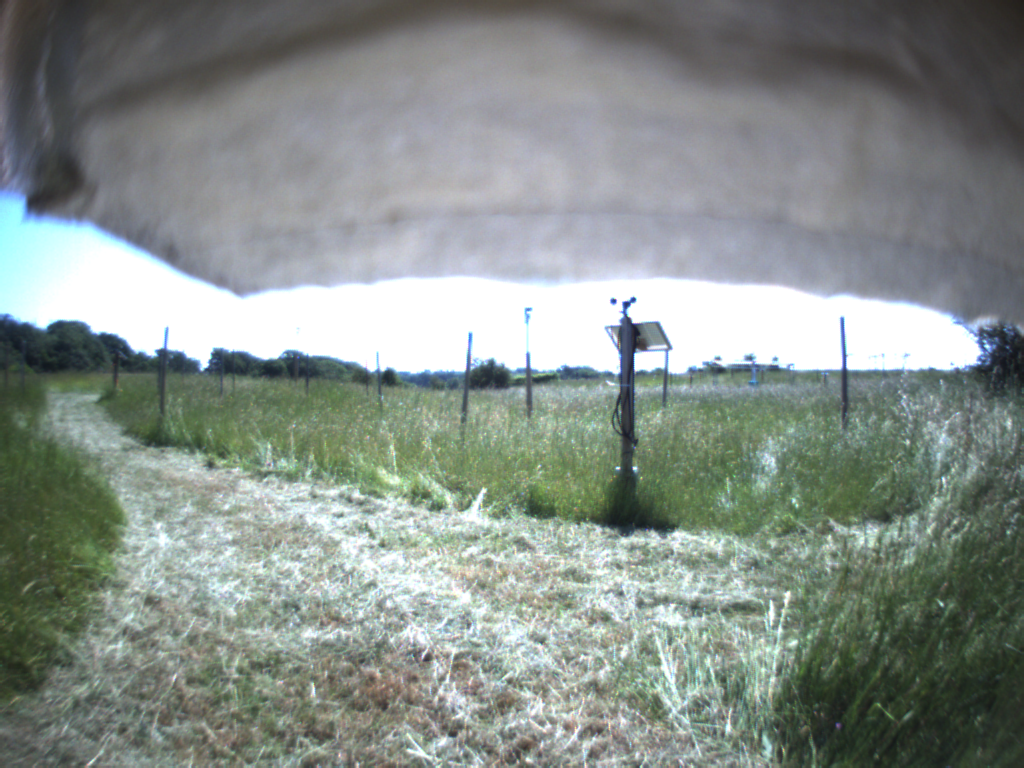}
\end{subfigure}
\begin{subfigure}{0.17\textwidth}
\includegraphics[width=\linewidth]{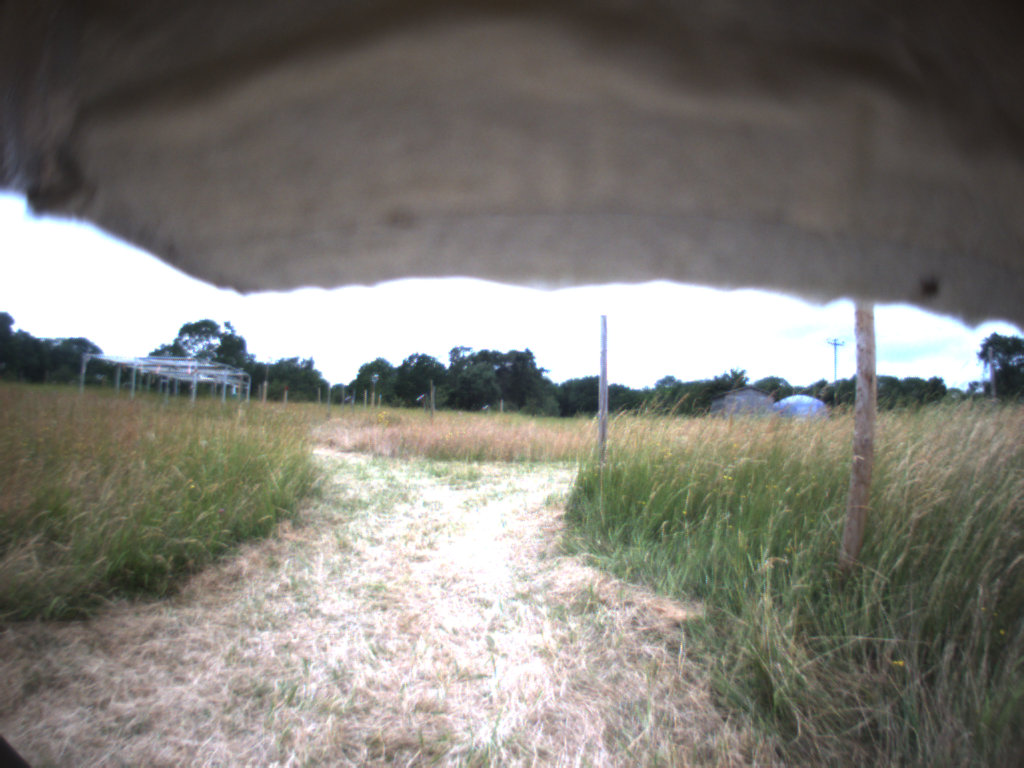}
\end{subfigure}
\begin{subfigure}{0.17\textwidth}
\includegraphics[width=\linewidth]{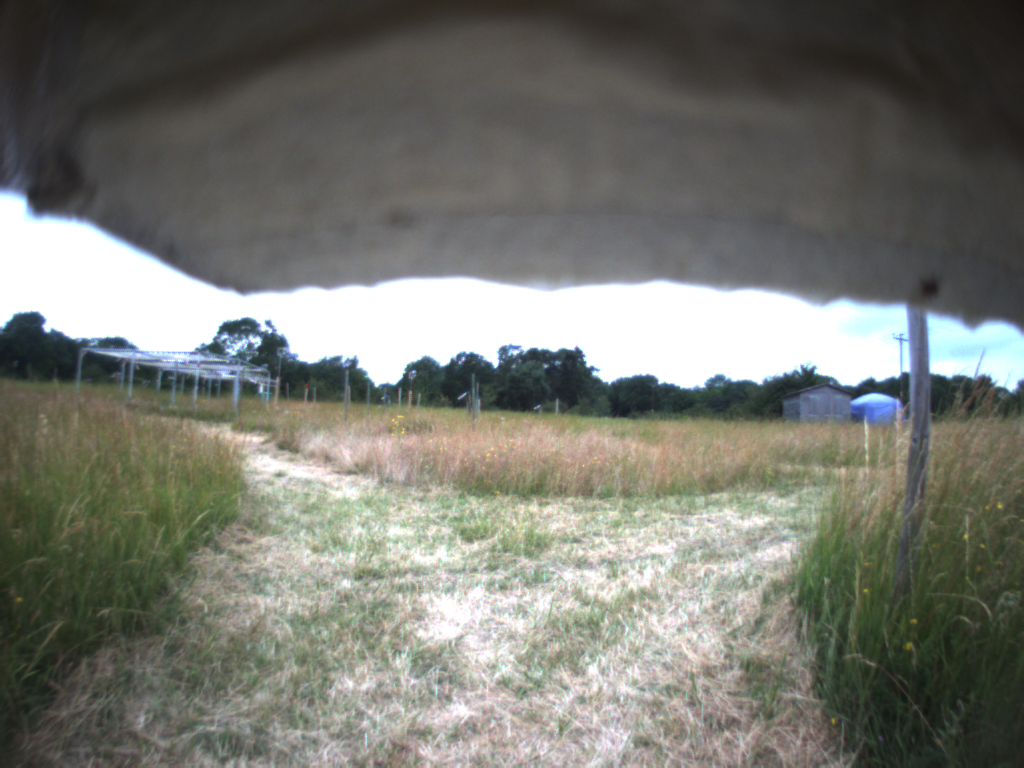}
\end{subfigure}
\begin{subfigure}{0.17\textwidth}
\includegraphics[width=\linewidth]{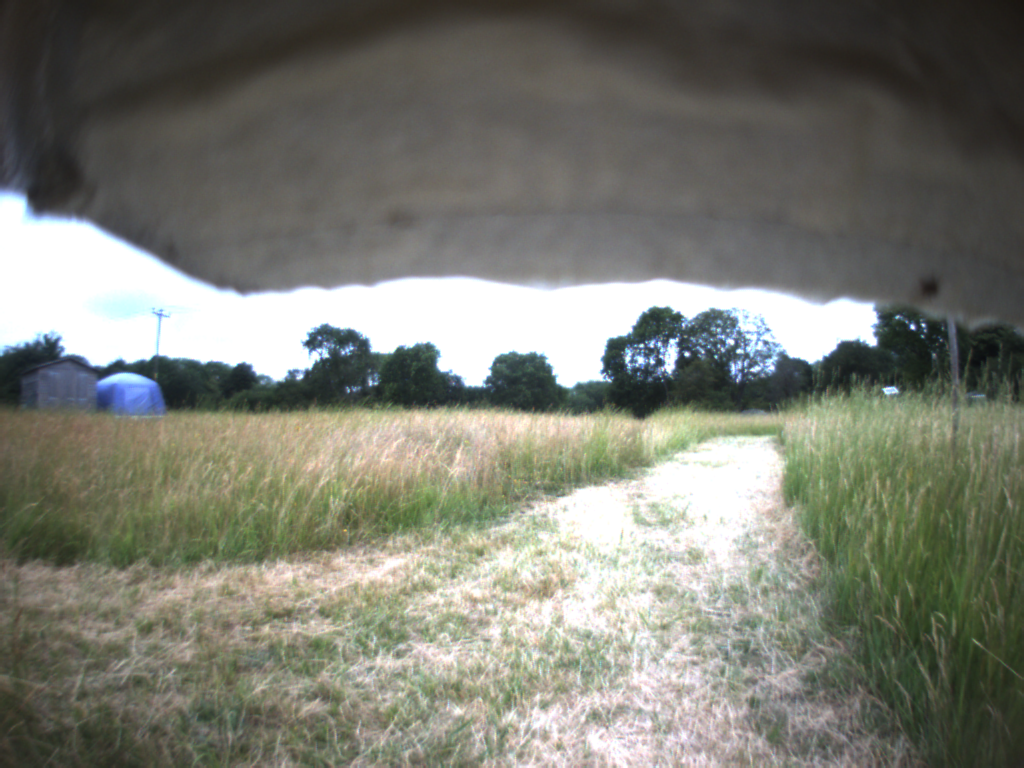}
\end{subfigure}
\begin{subfigure}{0.17\textwidth}
\includegraphics[width=\linewidth]{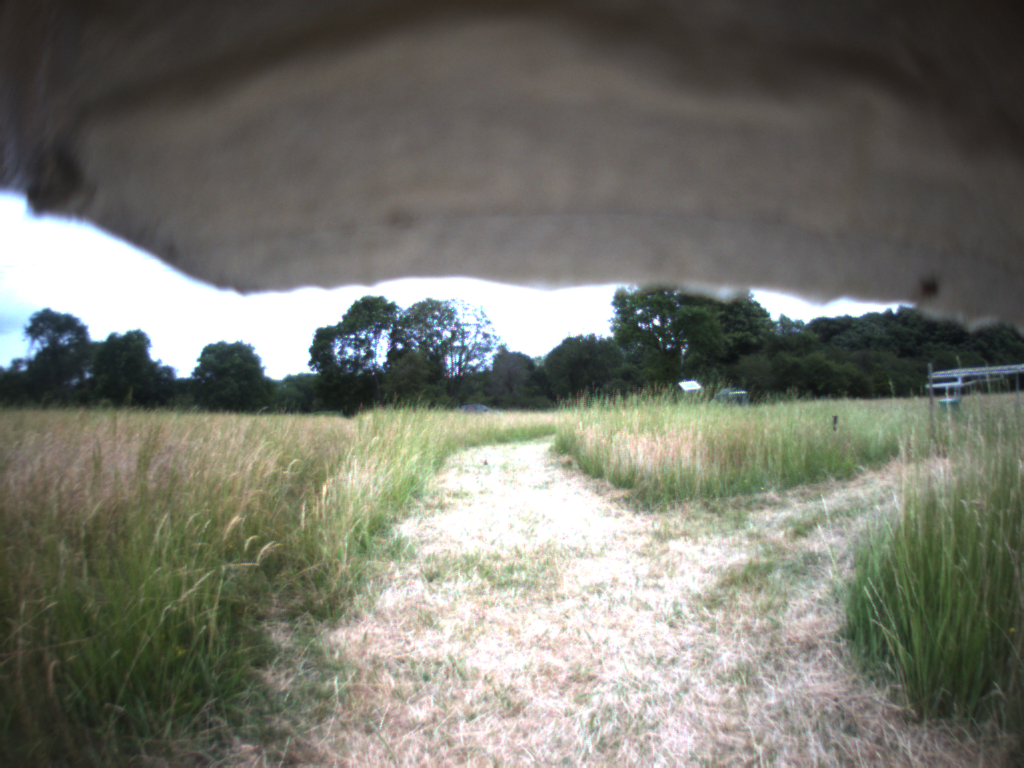}
\end{subfigure}
\begin{subfigure}{0.17\textwidth}
\includegraphics[width=\linewidth]{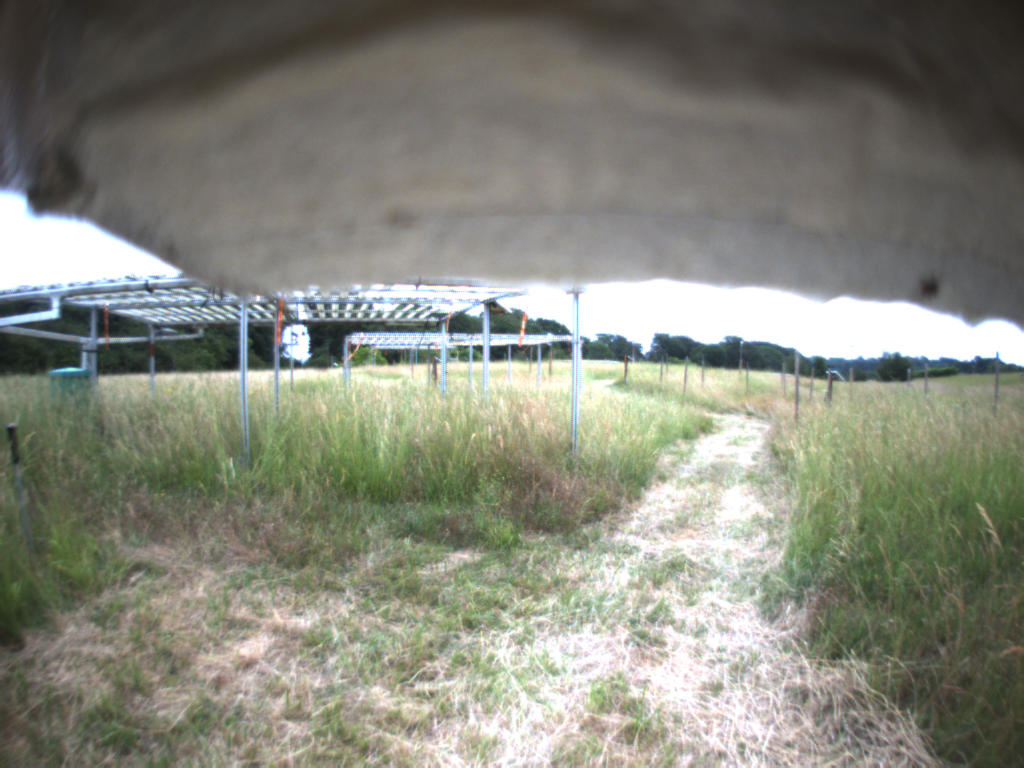}
\end{subfigure}
\caption{
\textbf{Sample Bumblebee2 images}.
All images are collected on the section of supergraph (\cref{sec:planning}) between places \texttt{B13} and \texttt{B15}.
\textit{Top Row}: Sequence collected on date \texttt{2023-06-14}.
\textit{Bottom Row}: Sequence collected on date \texttt{2023-06-28}.
}
\label{fig:bb2}
\vspace{-15pt}
\end{figure*}

For navigation (\cref{sec:dub4,sec:vtr}), the \textit{Bumblebee2} is used (\cref{fig:husky-in-field}).
As per~\cref{fig:bb2}, despite using an illumination invariant image transformation (\cref{fig:dub4}), we found sun glare to be problematic for feature matching to map frames, and that using a visor over the top of the camera -- despite being a very simplistic solution -- mitigated this issue and had the positive side-effect of ignoring features on clouds which would be ephemeral and distracting for our navigation system.

Alongside this, we equip the robot with thermal and multi-spectral cameras.
An example of the thermal camera is shown in~\cref{fig:thermal_frames}.
Here, thermal cameras capture infrared radiation emitted by plants, revealing temperature variations that indicate metabolic activity and stress levels.
Multi-spectral cameras analyse the absorption and reflection of different wavelengths by plants, providing insights into chlorophyll content, photosynthetic activity, overall health and estimate plant biodiversity \cite{jackson2022short}.

\section{Field Trials}

\subsection{Surface preparation}
\label{sec:mowing}

\Cref{fig:bb2} shows example images from the \textit{Bumblebee2} stereo vision camera mounted to the platform (\cref{fig:husky-in-field}). 
We have mowed the grass in between plant-growing experimental sites. 
This ground preparation was done to prevent spurious triggering of the safety curtain lasers (\cref{sec:scs}) by long blades of grass.
The mowing was carefully scheduled after the spring and early summer, during which the grass grows at its most rapid pace~\cite{jackson2024experimental} (to avoid having to mow the routes shown in~\cref{fig:hx_map} repeatedly).
Very practically, it was the obvious and convenient routes which the mowing vehicle could drive that dictated the exact topology of the ``supergraph'' (\cref{sec:planning}) used to enable autonomy around the site.

The second row in~\cref{fig:bb2}, from a later date, shows the grass in the distance (off the mowed path over which the robot is repeatedly driving) as significantly more yellow
This has implications for localisability, discussed in~\cref{sec:localisation_stats} below.

\subsection{Data collection \& mapping-localisation-mapping workflow}
\label{sec:mapping}

\begin{figure}
\centering
\includegraphics[width=\columnwidth]{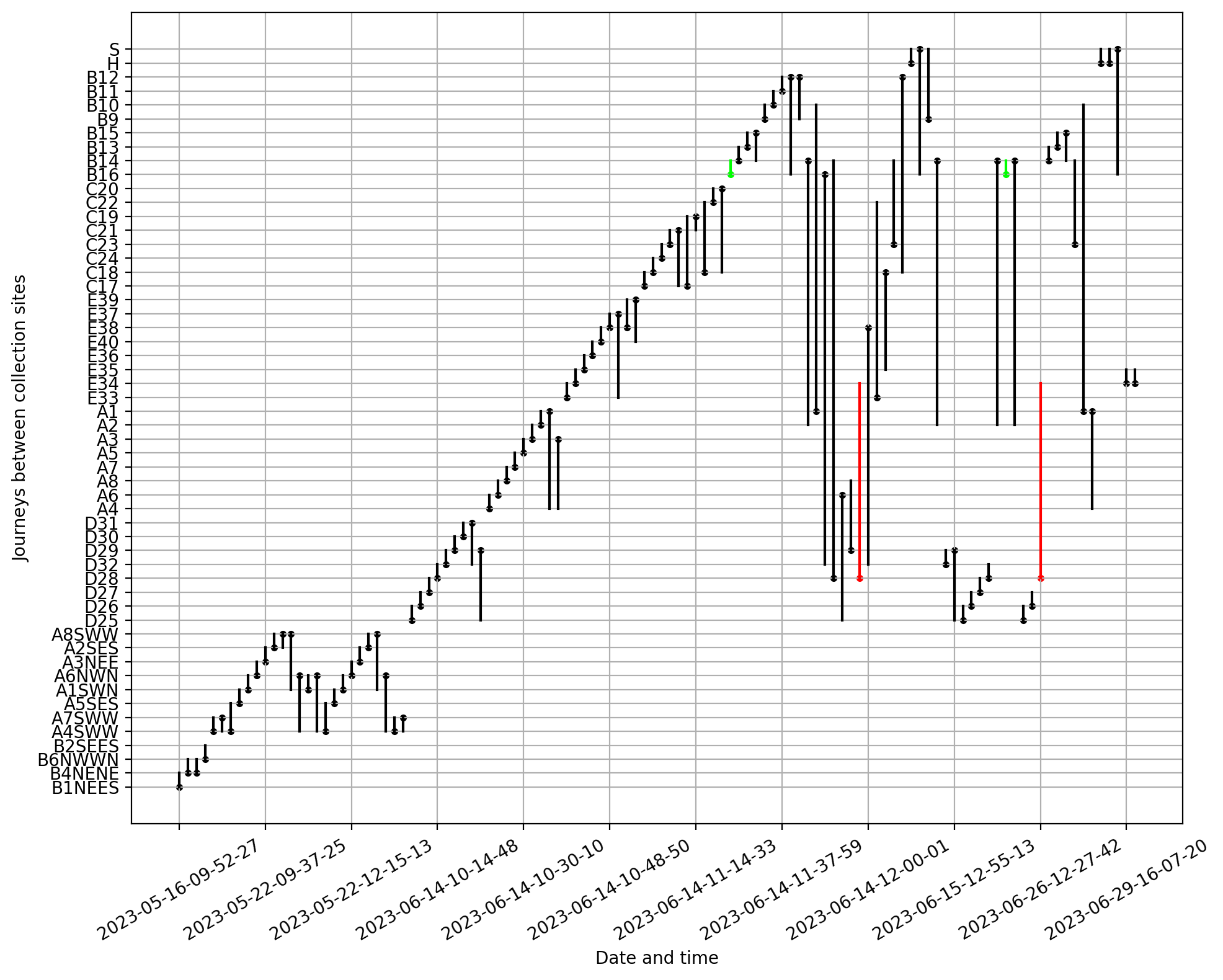}
\caption{
\textbf{Sequence collection timing diagram}.
Vertical lines indicate start/end places of collection.
The dots indicate the location at the start of the journey.
}
\label{fig:supergraph}
\vspace{-15pt}
\end{figure}

In~\cref{fig:supergraph} we illustrate our mapping activities.
Sequences are collected between plant growing data collection points, and then ``stitched'' together into the larger experience graph as well as supergraph, as mentioned in~\cref{sec:vtr}.
The black lines in~\cref{fig:supergraph} correspond to edges in the supergraph
(\cref{fig:supergraph,sec:planning}), containing many keyframes for visual localisation (\cref{sec:dub4}) recorded for each vertical black line.
Dots indicate the beginning of the journey.
The example edge \texttt{B16-B14} shown in~\cref{fig:hx_map} is highlighted green here, and has collection times along the $x$-axis of \texttt{2023-06-14-11-27-53} and \texttt{2023-06-26-10-19-22}.

Note that the initial set of nodes in the supergraph, or collection sites, is named differently from the rest (e.g. \texttt{B1NEES} versus \texttt{B9}).
Here, the map containing \texttt{A8SWW} to \texttt{B1NEES} was completely replaced and extended by the final map \texttt{S} to \texttt{D25}, with the initial map being used for initial trials in which we debugged our code and performed shakedown tests of the sensors and platforms.
Some interesting localisation experiments over this debugging map are nevertheless discussed in~\cref{sec:localisation_stats} and shown in~\cref{fig:repeat_logs_2023-05-16-13-02-00} below.

The majority of the sequences for the final map are collected on a single day, \texttt{2023-06-14}.
Due to the rapid withering of the grass (\cref{fig:bb2}) and its impact on localisability (\cref{sec:localisation_stats}), we needed to replace sequences at later dates (after which, localisation stabilised).
For example, the vertical red line at around \texttt{2023-06-15-12-55-13} between \texttt{E34} and \texttt{D28} has an earlier collection just before \texttt{2023-06-14-11-37-59} (also shown as a red line, for visibility).
This represents that sequence being replaced, but it is thereafter robust for localisation and autonomy, where in~\cref{fig:repeat_logs_2023-06-29-10-43-49,fig:repeat_logs_2023-06-29-14-20-34,fig:repeat_logs_2023-06-30-09-33-37,fig:repeat_logs_2023-06-30-11-50-16} below we see repeated autonomous traversals over this part of the supergraph (seen as the three long orange bars \texttt{D28-E34} about a third of the way through the journey).
This effect was not seen for all places and sequences across the map, reflecting that some paths (the ones replaced often) were more difficult to traverse smoothly and with good localisation.

Overall, there are $75$ pairs of places (edges in the supergraph,~\cref{fig:hx_map}) and thus $75$ sequences to collect.
After this initial collection of $75$ sequences, we needed to recollect sequences on $37$ more occasions.
However, $46$ of the initially collected edges in the supergraph (\SI{61}{\percent}) \textit{did not need to be recollected}.
Often (see~\cref{sec:lessons}), these replacements were required because, rather than localisability, not even care was taken in manually driving the robot around tight turns and with enough of a safe margin around obstacles, which was challenging for our simple feedback controller.

\subsection{Single- and multi-loop mission profiles}
\label{sec:timing_diagrams}

\begin{figure*}
\centering
\begin{subfigure}{0.9\textwidth}
\includegraphics[width=\linewidth]{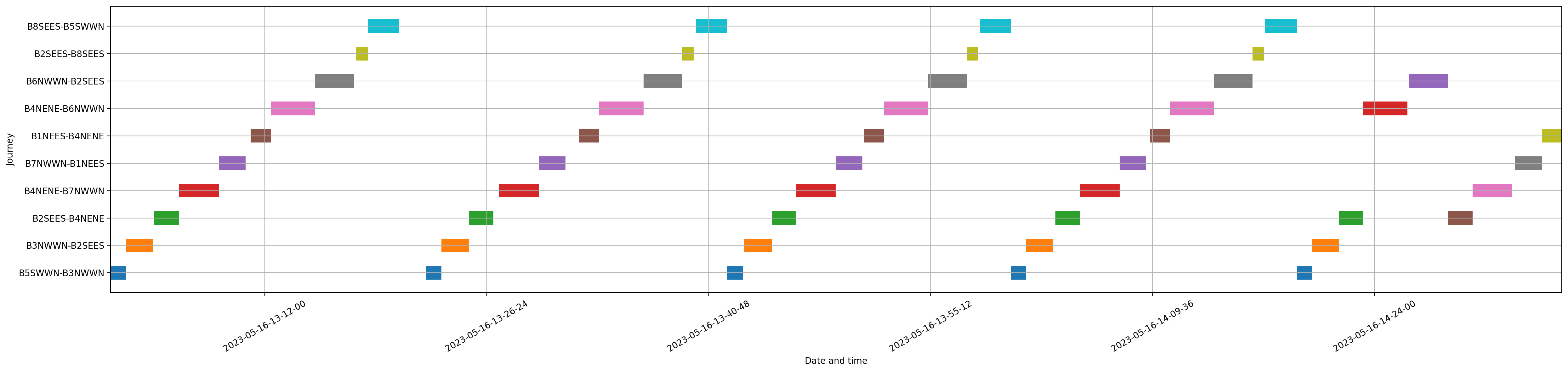}
\caption{\label{fig:repeat_logs_2023-05-16-13-02-00}}
\end{subfigure}
\begin{subfigure}{0.9\textwidth}
\includegraphics[width=\linewidth]{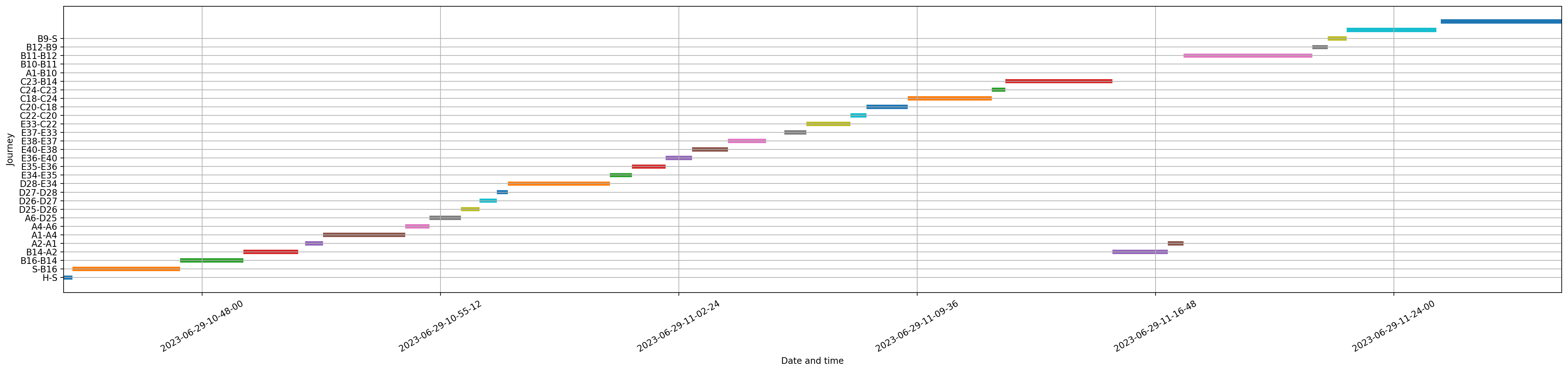}
\caption{\label{fig:repeat_logs_2023-06-29-10-43-49}}
\end{subfigure}
\begin{subfigure}{0.9\textwidth}
\includegraphics[width=\linewidth]{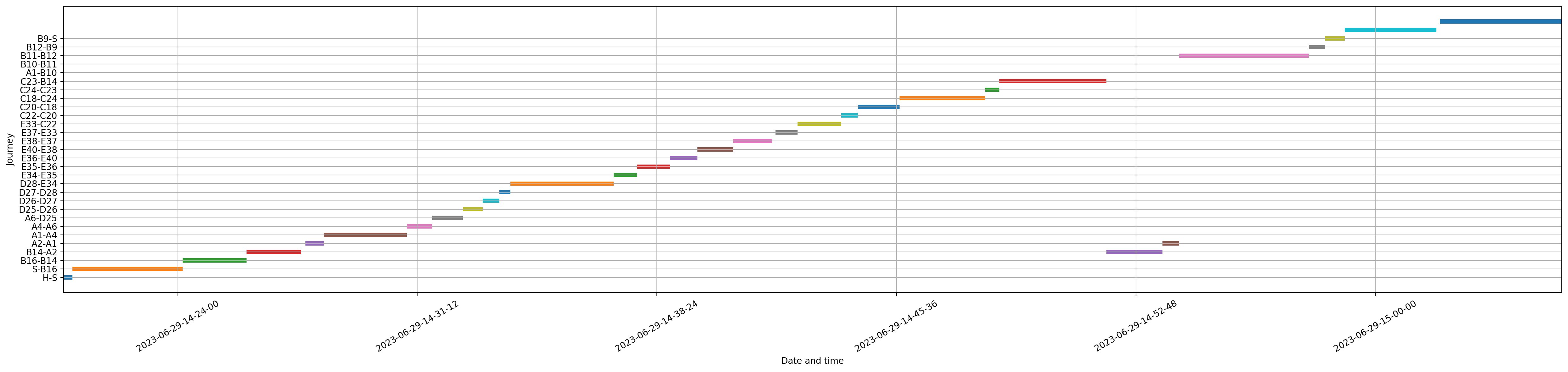}
\caption{\label{fig:repeat_logs_2023-06-29-14-20-34}}
\end{subfigure}
\begin{subfigure}{0.9\textwidth}
\includegraphics[width=\linewidth]{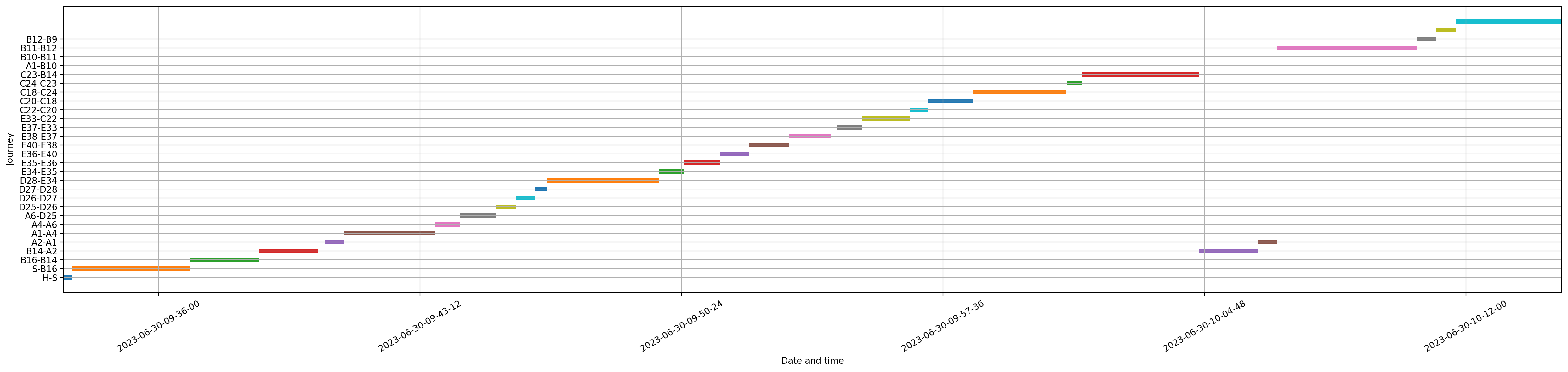}
\caption{\label{fig:repeat_logs_2023-06-30-09-33-37}}
\end{subfigure}
\begin{subfigure}{0.9\textwidth}
\includegraphics[width=\linewidth]{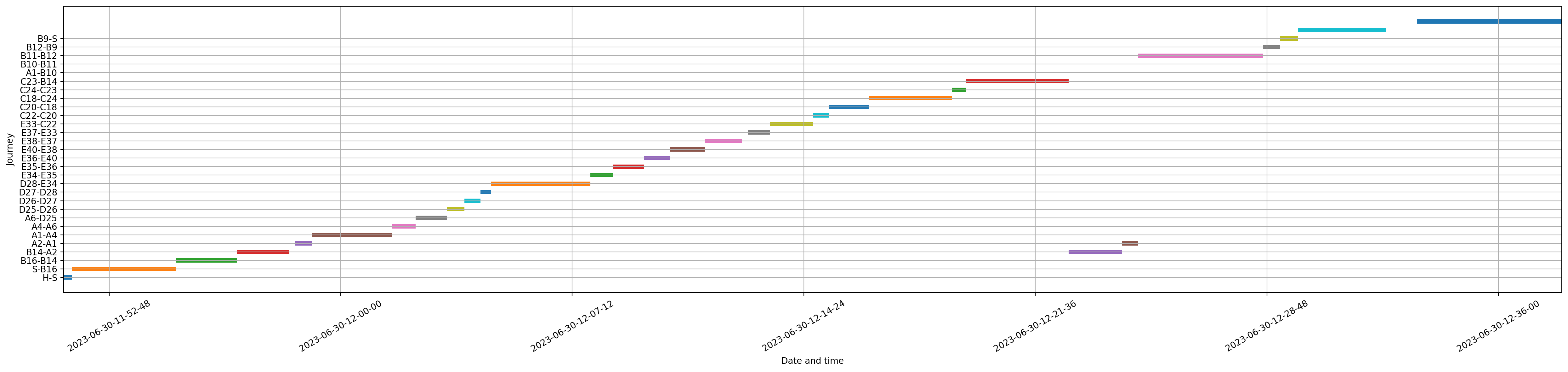}
\caption{\label{fig:repeat_logs_2023-06-30-11-50-16}}
\end{subfigure}
\caption{
\textbf{Timing diagrams} of autonomous mission.
The vertical axes show codes for experimental collection sites.
\subref{fig:repeat_logs_2023-05-16-13-02-00} is repeated over a different, earlier map to the other timing diagrams (for debugging), with different place codes, but is included here for interest.
}
\label{fig:repeat_logs}
\end{figure*}

\cref{fig:repeat_logs} show example timing diagrams for autonomous missions across multiple edges in the supergraph (across the site).

Each of these examples is in excess of \SI{1}{\hour} of autonomy, indicating the stability of our system.
In~\cref{fig:repeat_logs_2023-05-16-13-02-00} as discussed in~\cref{sec:mapping} we are using an earlier version of the map on a smaller subgraph (only places with codes \texttt{B}, i.e. the section of plant growing sites closest to the autonomous dock in~\cref{fig:hx_map}), but with many loops through that set of edges.

In~\cref{fig:repeat_logs_2023-06-30-09-33-37,fig:repeat_logs_2023-06-29-14-20-34,fig:repeat_logs_2023-06-29-10-43-49,fig:repeat_logs_2023-06-30-11-50-16}, over the final version of the map, the mission is much more ``linear'', as we aimed to traverse as much of the site as possible and visited each of the sections \texttt{A} through \texttt{E}.
To help with understanding the supergraph in~\cref{sec:planning,fig:hx_map}, consider traversals over the illustrated edges \texttt{S-16} and \texttt{B16-B14} which in~\cref{fig:repeat_logs_2023-06-30-09-33-37,fig:repeat_logs_2023-06-29-14-20-34,fig:repeat_logs_2023-06-29-10-43-49,fig:repeat_logs_2023-06-30-11-50-16} happen at the beginning of the journey (orange and green).

\subsection{$6$-week localisation statistics}
\label{sec:localisation_stats}

\begin{figure*}
\centering
\begin{subfigure}{0.96\textwidth}
\includegraphics[width=\linewidth]{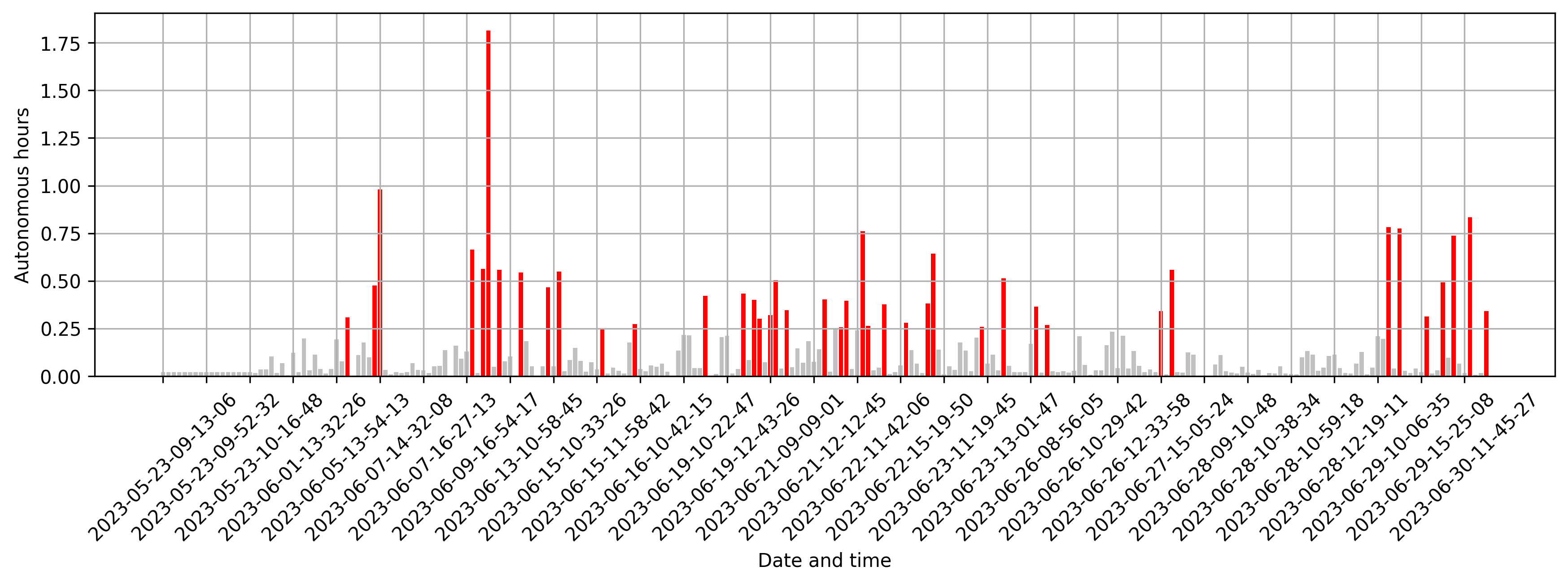}
\caption{\label{fig:autonomous_time}}
\end{subfigure}
\begin{subfigure}{0.96\textwidth}
\includegraphics[width=\linewidth]{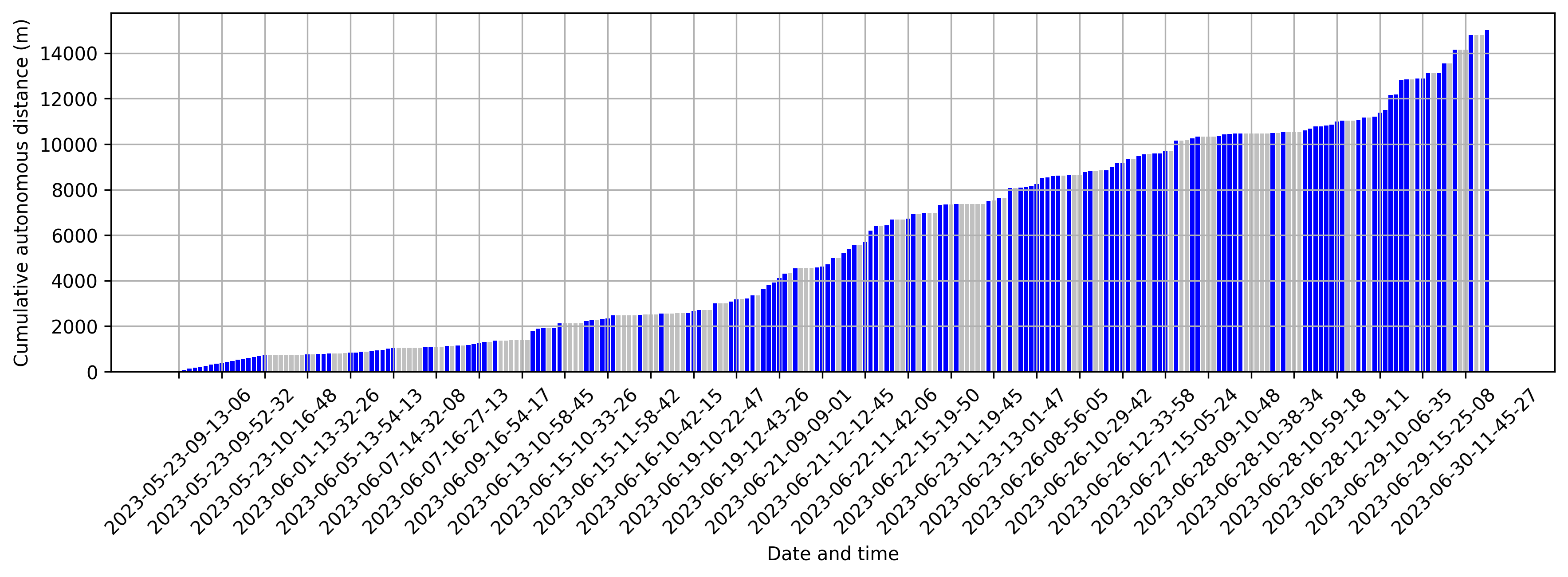}
\caption{\label{fig:odometer}}
\end{subfigure}
\begin{subfigure}{0.32\textwidth}
\includegraphics[width=\linewidth]{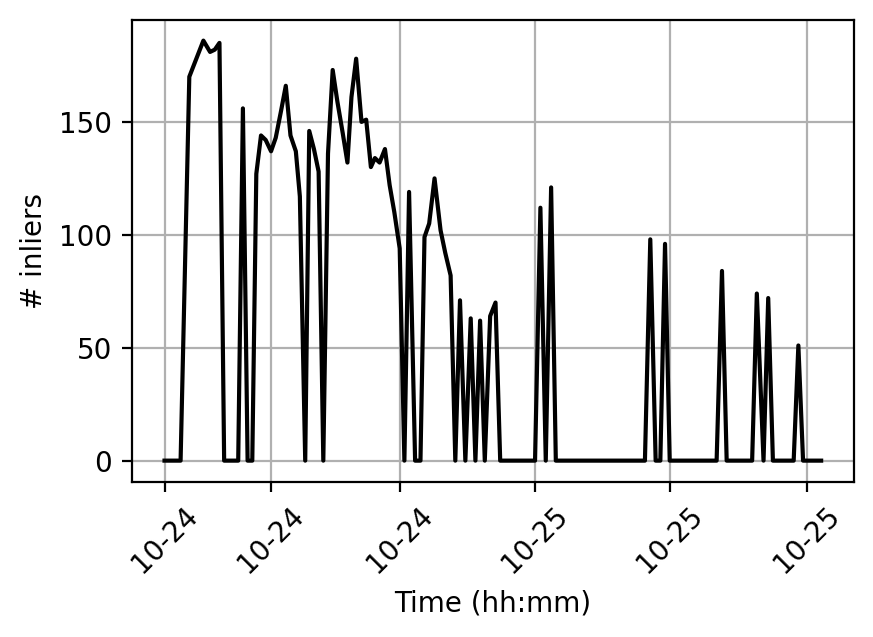}
\caption{\label{fig:num_inliers_2023-06-28-10-24-28}}
\end{subfigure}
\begin{subfigure}{0.32\textwidth}
\includegraphics[width=\linewidth]{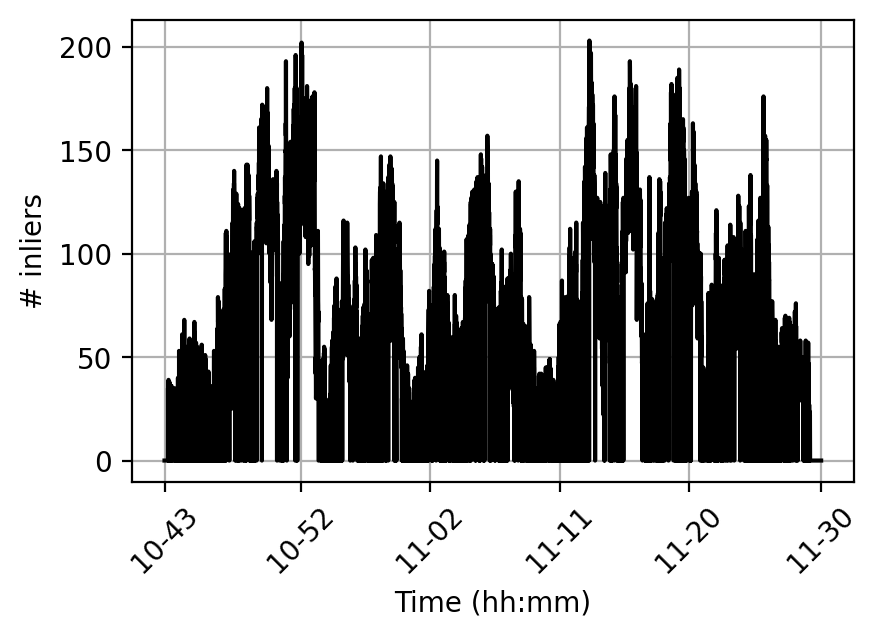}
\caption{\label{fig:num_inliers_2023-06-29-10-43-06}}
\end{subfigure}
\begin{subfigure}{0.32\textwidth}
\includegraphics[width=\linewidth]{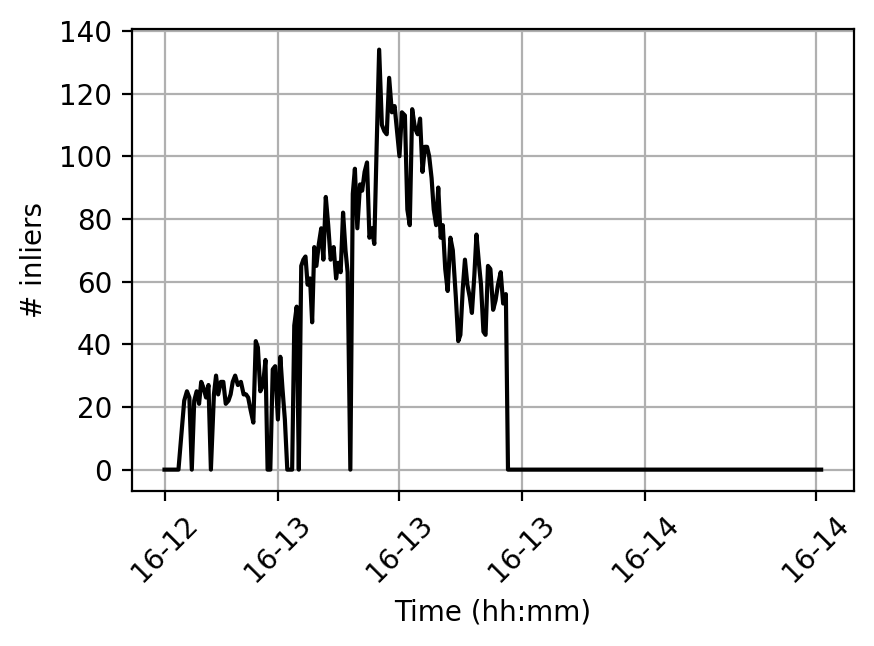}
\caption{\label{fig:num_inliers_2023-06-29-16-12-46}}
\end{subfigure}
\begin{subfigure}{0.32\textwidth}
\includegraphics[width=\linewidth]{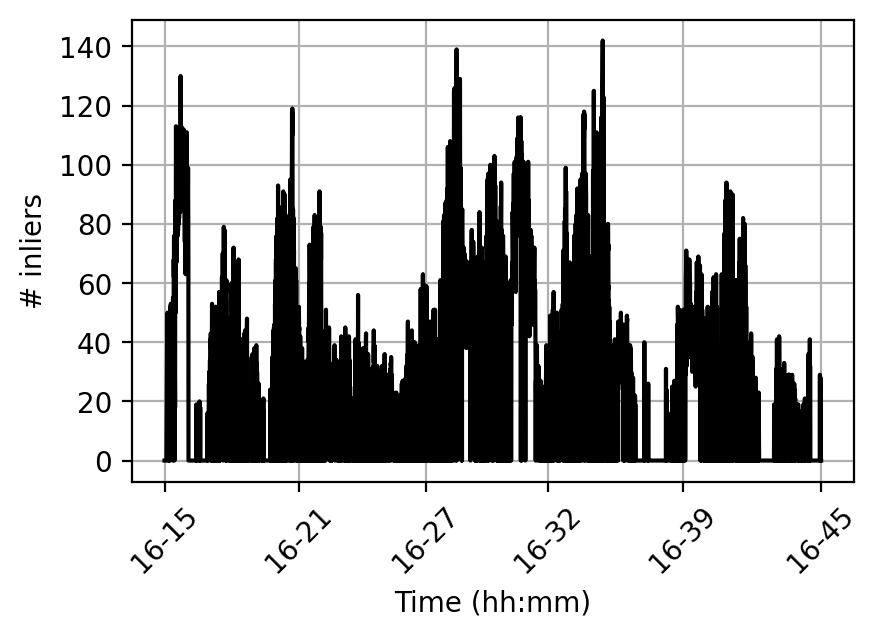}
\caption{\label{fig:num_inliers_2023-06-29-16-15-39}}
\end{subfigure}
\begin{subfigure}{0.32\textwidth}
\includegraphics[width=\linewidth]{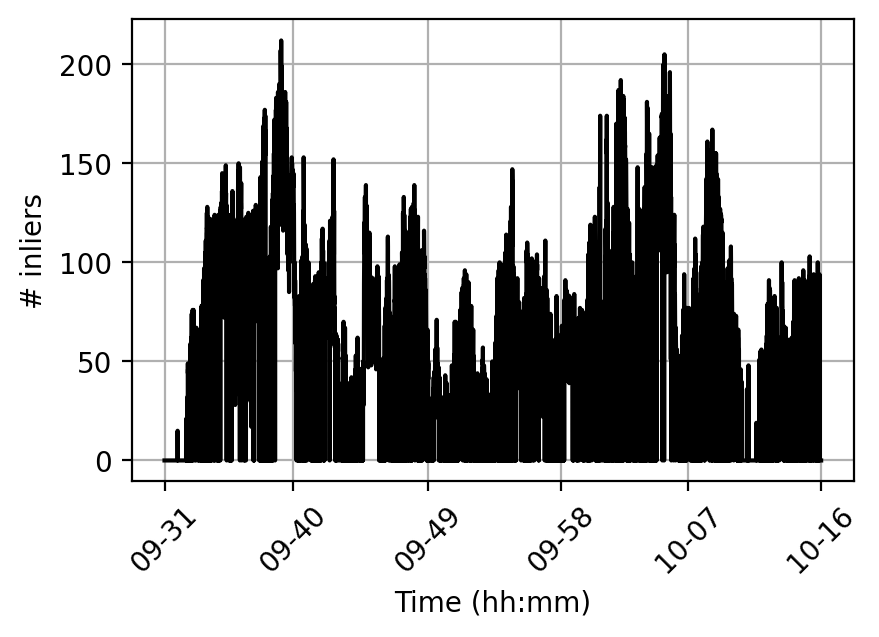}
\caption{\label{fig:num_inliers_2023-06-30-09-31-52}}
\end{subfigure}
\begin{subfigure}{0.32\textwidth}
\includegraphics[width=\linewidth]{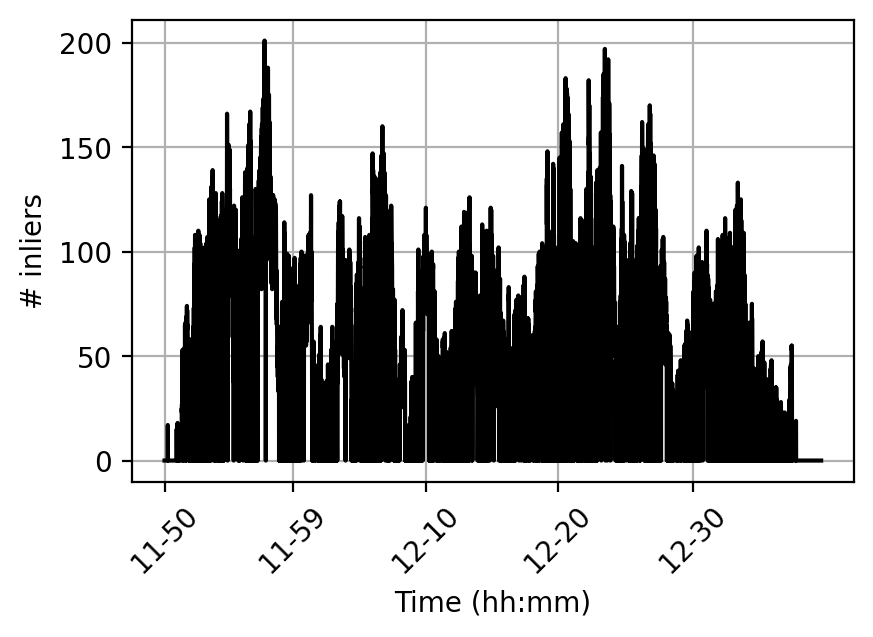}
\caption{\label{fig:num_inliers_2023-06-30-11-50-06}}
\end{subfigure}
\caption{
\textbf{Localisation statistics} from \texttt{2023-05-22} to \texttt{2023-06-30} (approximately \num{6} weeks).
}
\label{fig:localisation_success}
\end{figure*}

Beyond these specific examples in~\cref{fig:repeat_logs} of multi-stage, sitewide autonomous traversal, \cref{fig:localisation_success} shows statistics for localisation from \texttt{2023-05-22} to \texttt{2023-07-1} (almost \num{6} weeks of field trials, in total).

\cref{fig:autonomous_time} shows the total time while running autonomously.
Grey entries are those shorter than \SI{15}{\minute}, with red representing the longer autonomous runs.
This is often longer than \SI{45}{\minute}.
Our longest autonomous traversal was more than \SI{1}{\hour} and \SI{45}{\minute}.
As can be seen, our autonomous outings are punctuated by periods of grey entries as we fixed issues with the platform or the map data we had recorded (see \cref{fig:supergraph} for mapping activity), or needed to tune the controller, etc.

\cref{fig:odometer} shows the corresponding cumulative autonomous metres driven, such that by the end of our field trials, the robot had driven more than \SI{14}{\kilo\metre} autonomously.
Similarly to above, grey entries in this plot correspond to aborted autonomous trials that were shorter than \SI{10}{\metre}.

\cref{fig:num_inliers_2023-06-30-11-50-06,fig:num_inliers_2023-06-30-09-31-52,fig:num_inliers_2023-06-29-16-15-39,fig:num_inliers_2023-06-29-16-12-46,fig:num_inliers_2023-06-29-10-43-06,fig:num_inliers_2023-06-28-10-24-28} show the number of inlier feature matches from the live frame to the map frame localised against -- as examples, for particular missions. 
\cref{fig:num_inliers_2023-06-30-09-31-52,fig:num_inliers_2023-06-29-16-12-46,fig:num_inliers_2023-06-29-10-43-06,fig:num_inliers_2023-06-28-10-24-28} are examples of good missions, over extended periods (more than \SI{45}{\minute}).
Here, inliers from live frame to the map are available at all times, i.e. the robot remains localised and it never travels very far in open-loop and odometric drift does not cause it to become lost and abort autonomy.
\cref{fig:num_inliers_2023-06-29-16-15-39,fig:num_inliers_2023-06-30-11-50-06} are short trials corresponding to the grey entries in~\cref{fig:autonomous_time,fig:odometer}.
\cref{fig:num_inliers_2023-06-29-16-15-39} is an example of the robot becoming lost in this way, with the number of inliers dropping to and remaining at $0$.
A short (aborted) trial \cref{fig:num_inliers_2023-06-30-11-50-06} is included to show that the inliers regularly drop and stay at $0$, i.e. the robot only localises every few frames, sufficient to keep it under control.

\section{Lessons learned}
\label{sec:lessons}

In the course of preparing for and running these trials, we have learned the following key lessons:
\begin{itemize}
\item \textbf{Taught path complexity} often causes basic heading-only controllers to steer the robot off path, thus losing localisation. This is easy to confuse with localisation error itself, but smooth, simpler driving often fixes issues on a re-teach of a route section.
\item \textbf{Initialising localisation} in visually ambiguous environments is often tricky (after initialisation, though, localisation is often robust). The take away is to spend time building a usable manual initialisation GUI. 
\item \textbf{Battery management} can hamper debugging code and platform shakedown in the field.
Developing a quick charge and hot-swap routine is helpful.
\end{itemize}

\section{Conclusion}
\label{sec:conclusion}

We demonstrated our system in an extended field trial in a complex and highly dynamic ecosystem and showed that our pipeline is well suited to autonomous data collection for monitoring outdoor plant-growing experiments, with key implications not only for blue-sky research, but also for agricultural applications.
We intend to integrate the grass-friendly robotic platform presented in~\cite{kyberd2021hulk} into our future work due to its potential to address the sensitivity of the grassy surface we have been operating on, where conventional outdoor robots using skid steering pose a risk of damaging the surface.

%

\bibliographystyle{IEEEtran}
\bibliography{biblio}

\end{document}